\documentclass[lettersize,journal]{IEEEtran}
\usepackage{amsmath,amsfonts}
\usepackage{algorithmic}
\usepackage{algorithm}
\usepackage{array}
\usepackage[caption=false,font=normalsize,labelfont=sf,textfont=sf]{subfig}
\usepackage{textcomp}
\usepackage{stfloats}
\usepackage{url}
\usepackage{verbatim}
\usepackage{graphicx}
\usepackage{multirow}
\usepackage{multicol}
\usepackage{cite}
\begin{document}

\title{Text-guided 3D Human Motion Generation with Keyframe-based Parallel Skip Transformer}

\author{Zichen~Geng,
        Caren~Han, 
        Zeeshan~Hayder,
        Jian~Liu, 
        Mubarak~Shah,~\IEEEmembership{Fellow,~IEEE,} \\
        and~Ajmal~Mian,~\IEEEmembership{Senior Member,~IEEE}%
\thanks{Zichen Geng and Ajmal Mian are with the School of Electrical, Electronic and Computer Engineering, University of Western Australia, Perth, WA 6009 Australia (e-mails: zen.geng@research.uwa.edu.au; ajmal.mian@uwa.edu.au).}%
\thanks{Caren Han is with the University of Melbourne, Parkville, VIC 3010, Australia (e-mail: caren.han@unimelb.edu.au).}%
\thanks{Zeeshan Hayder is with the Commonwealth Scientific and Industrial Research Organisation (CSIRO), Canberra, ACT 2601, Australia (e-mail: zeeshan.hayder@data61.csiro.au).}%
\thanks{Jian Liu is with Hunan University, Changsha, Hunan 410082, China (e-mail: jianliu99@outlook.com).}%
\thanks{Mubarak Shah is with the University of Central Florida, Orlando, FL 32816, USA (e-mail: shah@crcv.ucf.edu).}%
}

\markboth{Journal of \LaTeX\ Class Files,~Vol.~14, No.~8, August~2021}%
{Shell \MakeLowercase{\textit{et al.}}: A Sample Article Using IEEEtran.cls for IEEE Journals}

\maketitle

\begin{abstract}
    Text-driven human motion generation is an emerging task in animation and humanoid robot design. 
    Existing algorithms directly generate the full sequence which is computationally expensive and prone to errors as it does not pay special attention to key poses, a process that has been the cornerstone of animation for decades. 
    We propose KeyMotion, that generates plausible human motion sequences corresponding to input text by first generating keyframes followed by in-filling. 
   We use a Variational Autoencoder (VAE) with Kullback–Leibler regularization to project the keyframes into a latent space to reduce dimensionality and further accelerate the subsequent diffusion process.      
    For the reverse diffusion, we propose a novel Parallel Skip Transformer that performs cross-modal attention between the keyframe latents and text condition.
    To complete the motion sequence, we propose a text-guided Transformer designed to perform motion-in-filling, ensuring the preservation of both fidelity and adherence to the physical constraints of human motion.
    Experiments show that our method achieves state-of-the-art results on the HumanML3D dataset outperforming others on all R-precision metrics and MultiModal Distance. KeyMotion also achieves competitive performance on the KIT dataset, achieving the best results on Top3 R-precision, FID, and Diversity metrics.
\end{abstract}


\section{Introduction} \label{sec:intro}

There is a growing interest in text-driven 3D human motion generation due to the rising demand for multimodal animations in the gaming and filming industries. This task can support designers in the creation of animations through straightforward text instructions, alleviating the need for labor-intensive designing and motion capture. Text-driven human motion generation is also useful for humanoid robot development, for generating human-like motions.

\begin{figure}[!ht]
    \centering
    \includegraphics[width=1\linewidth]{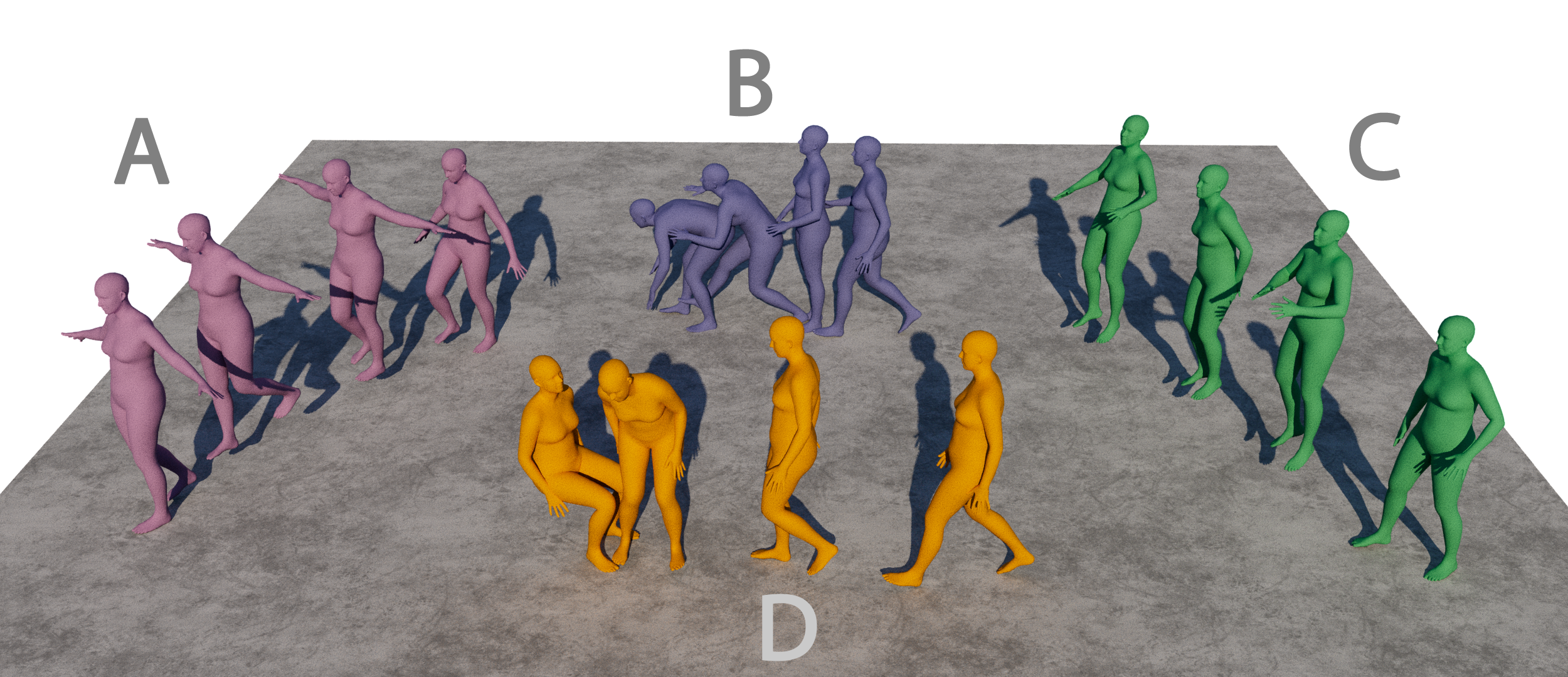}
    \caption{Generated keyframes by our method with input texts: (A) a person walks forward carefully placing one foot directly in front of the other; (B) a person walks forward, bends forward, walks backward; (C) a man takes his hands puts them on his hips and jumps up and down. (D) a person walks forward and then sits.}
    \label{fig:demo}
\end{figure}

Early methods for motion generation were based on autoencoders \cite{language2pose, hier}, VAEs \cite{a2m, ACTOR, humanise} and Generative Adversarial Networks (GANs) \cite{hpgan, bihmpgan, hmpgan, ahn2018text2action}. With the success of denoising diffusion models \cite{ddpmo, song2020score, ddpm} for image generation \cite{ddpm, scalableDIT} and video generation \cite{videodiff}, recent methods are adopting this technology for 3D human motion generation as well. However, existing methods apply generative algorithms directly on the whole motion sequences \cite{mofusion, motiondiffuse, temos}, which have significant redundancy, with mild continuous changes over time. Small interval motion sequences can be fully predicted from only the start and end pose. 
Learning relationships between all frames requires unnecessary computational resources and may also cause matching problems between text conditions and motion features.
To alleviate the computation problem, multi-stage models have been proposed \cite{t2mgpt, text2motion, mld} that learn lower-dimensional representations first and then perform motion generation in the low-dimensional space. 
While projecting to lower feature space benefits the conditional generation task, it may also lead to poor reconstruction results. Hence, a compromise between efficiency and reconstruction fidelity must be made. 

{We notice the fact that unsuitable keyframe selection would cause temporally non-uniform distribution or high proximity (collisions) between their representations i.e. the adjacent keyframes may be highly similar} This can be a potential problem for {both the infilling and} the generative algorithms which are assigned with the difficult task of removing noise. 
{Therefore, we propose a new keyframe selection method regarding both coherent frame collisions and uniform distribution.} 

For decades, the prevailing practice for animation design has been that an expert draws only the key poses (keyframes) which are then filled in by less experienced workers, or interpolated using quaternions in the case of 3D animation  \cite{keyframemanual}. To the best of our knowledge, such an approach has not been translated to 3D human motion generation in the existing literature. We argue that a keyframe-based approach is a good option for motion generation as well due to three reasons. (1) {It provides an additional dimension for data reduction without further compromising fidelity.} (2) {By applying proper keyframe selection strategies, the collision problem is alleviated to a great extent.} (3) The subsequent diffusion model will have a better chance of recovering the correct keyframes from noise given the large inter-frame differences of their latent representations.


In this paper, we propose KeyMotion, a 3D human motion generation method that first generates keyframes and then completes the full sequence. We use a VAE with Kullback–Leibler (KL) regularization to project keyframes to a latent space. We empirically show that the VAE learned from keyframes followed by in-filling leads to more accurate full-motion reconstruction compared to directly reconstructing the full sequence.  We propose a two-stream Parallel Skip Transformer (PST) based on cross-attention and skip connections that performs text-conditioned 3D human motion diffusion in the latent space learned by the VAE. By merging cross-modal features and avoiding information twisting, the proposed PST achieves improved denoising capabilities compared to individual parallel or skip Transformer architectures. 
This advantage is in addition to the fact that our PST is only recovering keyframes in latent space, which makes it extremely efficient and achieves higher accuracy given the uniqueness of keyframe latent representations. 
The denoised latent representation is then projected back to the 3D human motion space to recover the keyframes. To complete the sequence, we propose a motion Masked AutoEncoder (MMAE) based on a text-conditioned residual Transformer with masks on motion in-betweens. Our contributions are summarized as follows:


\vspace{-2mm}
\begin{itemize}
    \item We propose a text-guided motion generation method (KeyMotion) based on keyframes. This is the first method that {\em generates} full motion sequences by first generating keyframes, simplifying the complex and redundant full motion sequence representations without excessive information loss.

    \item We propose a Parallel Skip Transformer (PST) diffusion denoising model to generate consistent motion keyframes. PST effectively merges text and motion features by incorporating additional attention pairs. 
    
    \item We propose a motion Masked AutoEncoder (MMAE), a text-conditioned residual Transformer, to fill up in-between frames.
\end{itemize}

Experiments show that our method achieves state-of-the-art results on the HumanML3D dataset \cite{text2motion} outperforming all existing methods on the Top1, Top2, and Top3 R-precision metrics as well as MultiModal Distance. KeyMotion also achieves competitive performance on the KIT dataset \cite{KITML}, achieving the best results on Top3 R-precision, FID, and Diversity metrics. Our results show that generating keyframes followed by infilling outperforms existing full-sequence generation models. 
Our results also show that the VAE for keyframes reduces information loss i.e. when followed by the proposed MMAE, it produces more accurate full motion sequences compared to directly reconstructing the full sequence by other VAEs. Finally, our experiments show that our MMAE exploits the input text to recover faulty keyframes better than models without text guidance.

\section{Related Works}
\noindent {\bf Human Motion Generation:} Text-conditioned human motion generation aims to generate lifelike and high-fidelity motion sequences conditioned by text descriptions or instructions. 
Some works adopt RNN and its variants, while others employ Transformers as backbones for this task. Multiple text encoders have been used to extract text features as conditions to guide the generation, including ad-hoc GRU and pre-trained language models such as BERT \cite{BERT}, RoBERTa \cite{roberta}, and CLIP \cite{CLIP}. 
Most works for this task can be divided into one-stage and multi-stage methods. One-stage generative models learn the evidence lower bound of the raw motion data, while the multi-stage models first project the raw data into lower dimensional spaces and then apply generative algorithms. Examples of one-stage methods include  Conditional Variational Autoencoder (CVAE) based methods like TEMOS \cite{temos} and T2M \cite{text2motion}, and denoising diffusion models like MDM \cite{mdm}, MotionDiffuse \cite{motiondiffuse}, and FLAME \cite{kim2022flame}.  Multi-stage generative models include  TM2T \cite{tm2t} and T2MGPT \cite{t2mgpt} which are based on VQ-VAE \cite{VQVAE}, and  MLD \cite{mld} which is based on the latent diffusion model \cite{rombach2021highresolution}. TM2T \cite{tm2t} is the first model to apply VQ-VAE to capture motion clips as motion tokens. It employs GRUs \cite{gru} as a motion-language translation module to generate motion sequences. T2MGPT \cite{t2mgpt} further develops the VQ-VAE-based technique that uses GPT \cite{gpt} to generate motion token sequences. MLD \cite{mld} is the first model to apply the latent diffusion model to the text-driven motion generation task, achieving short inference time. However, the latent space projection of the full sequence using a VAE limits the upper bound of the motion fidelity. An overview of current approaches to human motion generation is given in Table \ref{tab:overview}.

\begin{table}[ht]
    \centering
    \caption{An overview of related works for text-driven human motion generation. GT stands for ground truth.}
    \begin{tabular}{r l l}
    \hline
    \hline
    Model & Approach & Motion Representation   \\
    \hline
    TEMOS\cite{temos} &  CVAE & full sequence \\
    T2M\cite{text2motion} &  CVAE & full sequence \\
    TM2T\cite{tm2t} &  VQ-VAE & motion token  \\
    MDM\cite{mdm} & diffusion & full sequence \\
    FLAME\cite{kim2022flame} & diffusion  & full sequence \\
    MotionDiffuse\cite{motiondiffuse} & diffusion & full sequence \\
    MLD\cite{mld} & latent diffusion & latent space \\
    DiffKFC\cite{diffkfc} & diffusion & full sequence \& GT keyframes\\
    T2MGPT\cite{t2mgpt} & VQVAE + GPT & motion token \\
    {\bf Ours} & latent diffusion & {\bf generated keyframes} \\
    \hline
    \end{tabular}

    \label{tab:overview}
\end{table}

\vspace{1mm}
\noindent \textbf{Diffusion Models:} Denoising Diffusion Probabilistic Models (DDPMs) \cite{ddpmo, ddpm} have demonstrated powerful capability in generative modeling across diverse domains including images, texts, and human motion sequences. Due to the large number of diffusion steps, the inference (denoising) time can be very long. Therefore, the Denoising Diffusion Implicit Model (DDIM) \cite{ddim} is proposed that skips part of the Markov chain during the reverse diffusion process. This reduces the number of time steps but the dimensionality of the original data still limits the generation speed. Rombach et al. proposed the Latent Diffusion Model (LDM) \cite{lmd} to accelerate the diffusion process by reducing the sample dimensionality. 
Peebles et al \cite{scalableDIT} proposed a scalable Transformer architecture to replace the U-net denoiser employed by LDM and achieved better performance on the text-to-image task. As the first model applying LDM to the text-to-motion task, MLD \cite{mld} employs a VAE with KL regularization to learn a latent representation and utilizes a skip Transformer denoiser to execute the reverse diffusion process. However, in practice, a trade-off between dimensionality reduction and denoising accuracy must be made. Reducing the latent dimension makes it more efficient for the denoiser to restore the latent samples. However, VAE-based latent space encoding/sampling of the full sequence limits the reconstruction fidelity and makes it challenging for the denoiser to recover the sample. 
Nevertheless, compared to diffusion-based models like MotionDiffuse \cite{motiondiffuse} and MDM \cite{mdm}, the latent diffusion model is a promising approach in motion generation due to its high inference efficiency and has also shown promise for motion prediction \cite{belfussion}. 
    
\vspace{1mm}
\noindent \textbf{Motion Infilling} 
aims to fill up frames in an interval between given frames. 
While \cite{motioninbetween2stage, conditionalmotioninbetween, robustmotioninfilling} use the start and end frames to fill in the gaps, \cite{rnnhumandynamic, trajhmp, dang2021msr, onhmprnn} use a start sequence to predict the subsequent frames.
Motion in-filling is traditionally achieved by interpolating between adjacent keyframes through deterministic algorithms like linear interpolation (LERP) or spherical linear interpolation (SLERP) \cite{slerp}. 
These algorithms have been foundational in the development of deep learning-based methods that utilize diverse architectures for infilling. Examples include CNNs \cite{cnn-infilling}, recurrent neural networks \cite{unifiedcvae} (RNNs, LSTMs, GRUs), and Transformers \cite{unifiedtransformer, continuousinterpolation, posegpt}.
In contrast to these approaches, the proposed Motion Masked AutoEncoder goes a step further by incorporating input text to guide the infilling, providing a mechanism to rectify errors stemming from faulty keyframes.


\begin{figure*}
    \centering
    \vspace{-3mm}
    \includegraphics[width=\textwidth, trim={0 18.5cm 0 0.5cm}, clip]{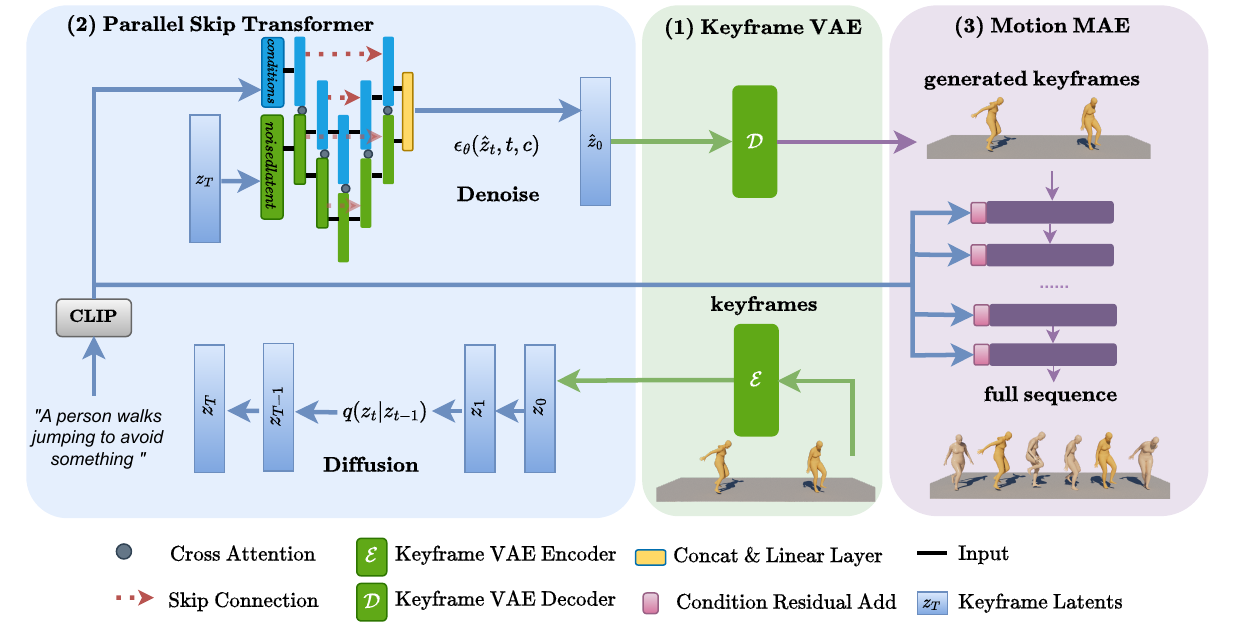}
    \vspace{-4mm}
    \caption{KeyMotion illustration. (1) Keyframe VAE encodes ($\mathcal{E}$) / decodes ($\mathcal{D}$) the keyframes in latent space. Keyframes in latent space are diffused to Gaussian noise. (2) Parallel Skip Transformer performs reverse diffusion in latent space using input text as conditioning. Denoised latent keyframes are decoded back to human motion space. (3) MMAE, a text-guided Transformer, performs keyframe infilling to produce the full sequence.}
    \label{fig:keymotion}
    \vspace{-4mm}
\end{figure*}

\section{KeyMotion Method}

Figure \ref{fig:keymotion} shows an overview of the proposed KeyMotion which comprises three stages; 
latent space projection, keyframe generation, and motion in-filling. 
The first step involves the meticulous selection of appropriate keyframes. \textbf{
}Since 
the frame representation is redundant and for reasons discussed in Sec~\ref{sec:intro}, we learn a VAE to project only the keyframes and their indices to a low dimensional latent space. Diffusion is performed in the latent space with PST and the denoised latent is decoded by the VAE to get the keyframes and their indices. MMAE then infills the missing frames to complete the motion sequence.

\subsection{Keyframe Selection}
Various definitions exist for keyframes. Some works define them as the frames that allow the remainders to be easily interpolated \cite{roberts2018optimal}, while others regard them as the frames that approximate the whole sequence the most. To study the performance of keyframe reconstruction and generation abilities for different keyframe selections, temporally uniform frames serve as the baseline. In practice, we found that the keyframe selection by minimizing interpolation error \cite{roberts2018optimal} does not explicitly solve the problem of close proximity. Hence, we propose a new method to select the keyframes to explicitly solve this issue. We define keyframes as a sequence along the path with the largest accumulated distance between adjacent keyframes.

\vspace{-2mm}
\begin{align}
D_{i,j}^{m} = \max_{k} \left( D^{m-1}_{i,k} + d_{k,j} \right),
\end{align}

\noindent where the $D_{i,j}^m$ denotes the maximum accumulated distance between each adjacent frames from $i$-th frame to $j$-th by picking $m$ frames in the interval, the $d_{k,j}$ denotes the (Euclidean) distance of motion features from $k$-th to $j$-th frame. This problem could be definitively solved using dynamic programming with a time complexity of $O(N^3)$, where $N$ is the full sequence length.
Compared with the minimum interpolation error optimal method and uniform selection, the Maximum Distance Optimal method (MDO) can maintain more information, produce more uniformly located frames in the original sequences, and avoid their close proximity.

\vspace{-3mm}
\subsection{Keyframe VAE (Variational Autoencoder)} \label{sec:VAE}

The keyframe training process involves the utilization of a variational autoencoder (VAE) with KL divergence regularization, aiming to learn the latent space corresponding to the keyframes. {Recall that the keyframe position must also be learned by the VAE, otherwise this information would not be available during inference. For this, we concatenate the position information with the keyframe before passing to the VAE during training.}
Specifically, given a keyframe sequence $F = [f_1, f_2, \cdots, f_K]$ with $f_k \in \mathbb{R}^d$, {and the corresponding keyframe position indices $P = [p_1, p_2, \cdots, p_K]$, with $p_k \in \mathbb{N}^1$}, where $K$ is the number of keyframes and $d$ is the dimension of each frame, we map the sequence and its position into a latent space $\varPhi = \{\varphi \}, \varphi\in\mathbb{R}^{l\times d_l}$ where $l$ is the latent size and $d_l$ the dimensionality. Specifically, we jointly train an unconditional encoder $\mathcal{E}$ and a latent decoder $\mathcal{D}$, for latent space projection and reconstruction of the keyframes, respectively. 
We employ Transformers for encoding/decoding. The keyframes are concatenated with $s_l$ learnable latent tokens, and the resulting composite is then processed through an encoder $\mathcal{E}$ augmented by sinusoidal positional encoding (SPE) \cite{vaswani2017attention}. This process yields the latent mean $\mu_l$ and standard deviation $\sigma_l$. Subsequently, the decoder $\mathcal{D}$ employs an SPE sequence as the query and the reparameterized latent variable as both the key and value for the reconstruction of the keyframes.

Note that the forward kinematics method cannot be used to directly obtain joint positions because the frame representation, while redundant, lacks explicit global joints. Furthermore, precise joint positions cannot be ensured by a straightforward MSE loss of the motion representation. Hence, we apply the constant bone length constraint during training to better control the VAE's reconstruction quality. This improves the consistency of the relative joint positions.

The objective function for the VAE training is a combination of reconstruction loss, KL divergence, bone length constraint, and position loss
\vspace{-2mm}
\begin{equation}
\mathcal{L}_{vae} = \mathcal{L}_{rec} + \lambda_{kl}\mathcal{L}_{kl} + \lambda_{bl}\mathcal{L}_{bl} + \lambda_{pos}\mathcal{L}_{pos}, 
\end{equation}
where $\mathcal{L}_{rec}$ denotes the reconstruction loss and is formulated as follows
\begin{equation}
\mathcal{L}_{rec} = \frac{1}{K}\sum_{k=1}^{K}||X_k - \hat{X}_k||_2.
\end{equation}
The $\mathcal{L}_{kl}$ is the KL divergence {with a Gaussian Distribution reference} and $\lambda_{kl}$ is the coefficient to control the relative importance of this loss as follows:
\begin{equation}
\mathcal{L}_{kl}=\frac12\sum_{i=1}^{d_l} (\sigma_i^2 + \mu_i^2 -1 -\log{\sigma^2_i}), 
\end{equation}
where $\sigma_i$ is $i$-th standard deviation of the latent variables, and $\mu_i$ is the mean value. Finally, loss $\mathcal{L}_{bl}$ enforces the constant bone length constraint, with $\lambda_{bl}$ as a coefficient to control the relative importance of the loss. 
\begin{equation}
\mathcal{L}_{bl} = \frac{1}{K}\sum_{k=1}^{K}\sum_{i=1}^{J} ||||j^{(i)}_k - j^{(i,p)}_k||_2 - ||\hat{j}^{(i)}_k - \hat{j}^{(i,p)}_k||_2 ||_2,
\end{equation}
where $j^{(i)}_k$, $\hat{j}^{(i)}_k$ are the absolute positions of the $i$-th ground truth / generated joints in the $k$-th frame, $j^{(i,p)}_k$, $\hat{j}^{(i,p)}_k$ are their parent joint, $K$ is the number of keyframes, and $J$ is the total number of joints.

\begin{equation}
\mathcal{L}_{pos} = \frac{1}{K}\sum_{k=1}^{K} |p_i - \hat{p}_i|,
\end{equation}
where $p_i$ is the position index of the $i$-th keyframe, and $\lambda_{pos}$ is the coefficient of reconstruction of the exact position index. 

\subsection{Parallel Skip Transformer Denoiser}

The VAE is employed to project all keyframes and corresponding position indices into a latent space for subsequent diffusion. To facilitate reverse diffusion (denoising), we introduce a two-stream Parallel Skip Transformer (PST), effectively integrating text and motion features through the inclusion of additional attention pairs.

Existing models~\cite{mld, mdm, motiondiffuse} typically involve a straightforward concatenation of pooler outputs from the CLIP model \cite{CLIP} with latent variables, followed by feeding them into a single-stream Transformer. In contrast, our proposed Parallel Skip Transformer (PST) employs two parallel streams with skip connections, resulting in an enhanced performance compared to the utilization of either parallel or skip Transformer alone.


\begin{figure}[t]
  \centering
    \includegraphics[width=1\linewidth, trim={0.5cm 18.5cm 4cm 1.2cm }, clip]{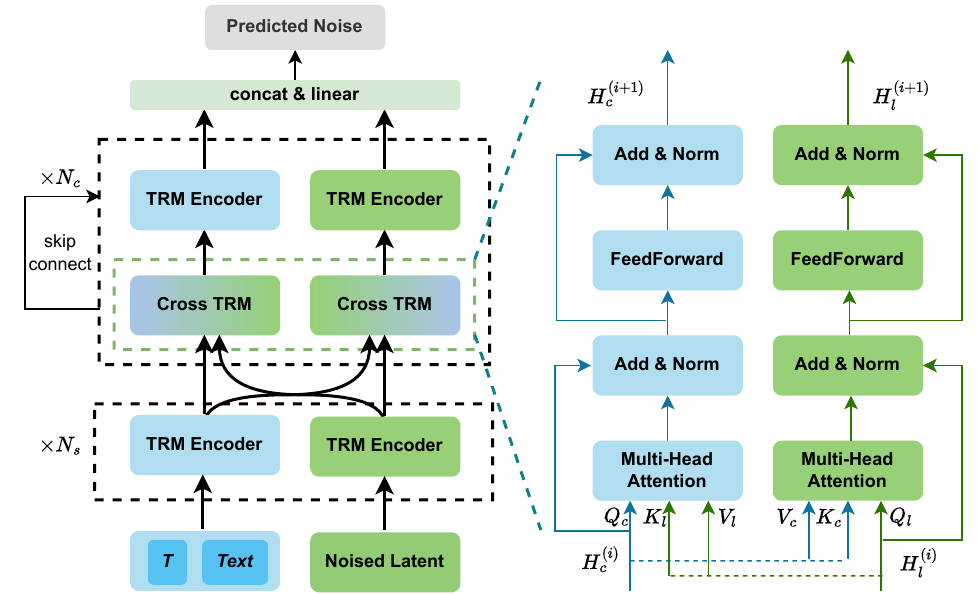}
  \caption{Parallel Skip Transformer and two-stream cross attention Transformer layer for denoiser module. By crossing attention operations between the text condition and latent variables, our model learns more stable textual information. $T$ is the timestep of the diffusion process, TRM is the basic Transformer, $H_c^{(i)}, H_l^{(i)}$ is the condition and latent hidden space through $i$-th cross attention Transformer.}
  \vspace{-3mm}
  \label{fig:cotrm}
\end{figure}

The proposed PST is inspired by ViLBERT \cite{vilbert} which uses a two-stream parallel Transformer (but no skip connections) for learning visual-linguistic representations. 
In Figure \ref{fig:cotrm}, the architecture of the proposed Parallel Skip Transformer (PST) is illustrated. This design aims to enhance the denoising capability in a text-conditioned manner by effectively combining text and keyframe features, mitigating information twists associated with multilayer cross-attention through the incorporation of skip connections.
{The text and time step representations are concatenated as a condition and passed to $N_s$ layers of the Transformer encoder, and then through $N_c$ parallel cross-Transformer layers. The $i$-th layer of the cross Transformer is skip-connected with the $(N_c-i)$-th layer by first concatenation and then projection via an MLP. The two-stream cross-modal representation after the cross Transformer is concatenated and projected to the latent dimension.} 

We use a scaled-linear noise scheduling method to add noise to the keyframe latents for each diffusion time step. Our denoiser aims to predict the added noise in the latent variables for each diffusion step using the objective function:
\begin{equation}
    L_{dn} = \mathbb{E}_{\epsilon, z, t, c}  ||\epsilon - \epsilon_{\theta}(z_t,t,c)||_2, 
    \label{eq:latent}
\end{equation}
where the $\epsilon_{\theta}$ predicts noise using the learnable parameter $\theta$, the $z$ is the keyframe latent variable, $t$ is the diffusion timestep, and $c$ is the text condition. We apply the classifier-free guidance in the training and inference phases. The predicted noise in the inference phase is given by:
\begin{align}
\epsilon_{\theta}(z_t, t, c) = s \epsilon_{\theta}(z_t, t, c) + (1-s) \epsilon_{\theta}(z_t, t, \emptyset), 
\end{align}
where $s$ is the classifier-free guidance factor and $\emptyset$ is the unconditional token. In the training phase, we randomly set the text conditions to be unconditional tokens with a proportion $p_u$ and use the loss function in Eq.~\ref{eq:latent} {to estimate the displacement between the diffusion noise and the predicted noise by $\epsilon_\theta$.}

\subsection{Motion Masked AutoEncoder}

The proposed Motion Masked AutoEncoder (MMAE) is a text-conditioned Transformer. Unlike other motion-in-filling models which only take motion sequence as input, we add the text condition to the infilling model as well to enhance its performance. We found that the text condition gives weak guidance when directly concatenated with the keyframes. Therefore, we propose a conditional residual Transformer by adding CLIP text representations to the leading position in each layer of the Transformer (see Fig.~\ref{fig:keymotion}). We discovered that for the motion in-filling with the keyframe, simple MSE loss on the redundant representations or joint position causes discontinuity. To solve this issue, we add the constant bone length constraint together with the velocity smoothness constraint to the loss function. The complete loss for MMAE is given by:

\begin{small}
\begin{align}
    \mathcal{L}_{MMAE} &= \mathcal{L}_{rec}  + \lambda_{fk}\mathcal{L}_{fk} + \lambda_{bl}\mathcal{L}_{bl} +  \lambda_{sm}\mathcal{L}_{sm}, \\
    {\rm where} ~~ \mathcal{L}_{fk} &= \frac{1}{N}\sum_{i=1}^N||\text{FK}(X_i)-\text{FK}(\hat{X}_i)||_2, \\
     ~~~\mathcal{L}_{sm} &= \frac{1}{N-1}\frac{1}{J} \sum_{i=1}^{N-1}\sum_{k=1}^{J}||\dot{j}_{i}^{k}-\hat{\dot{j}}_{i}^{k}||.
\end{align}
\end{small}
In the above, FK is the forward kinematic function converting the redundant representation to joint positions, and the $\dot{j}$ is the joint velocity. Inspired by Masked Autoencoder \cite{MAE}, we mask the missing frames in between by zeros and force the MMAE to predict the exact motion representation of both the keyframes and the in-between frames i.e. the full sequence. For the inference, there is no ground truth keyframe position available, and we use the generated keyframe indices by the VAE decoder to place the keyframes in the corresponding positions of the full sequence. 

\section{Experiments}

We perform comparative experiments on two benchmark datasets. 
 
\vspace{1mm}
\noindent \textbf{Datasets}. We use two popular benchmark datasets. 
(1) HumanML3D \cite{text2motion}, which is the largest published dataset that annotates the
existing 3D motion capture datasets AMASS \cite{AMASS} and HumanAct12 \cite{a2m}, containing 14,616 motion sequences with 44,970 textual descriptions. Each frame has a redundant representation of the concatenation of root velocities, local joint positions, joint velocities
6-D joint rotations, and foot contact labels. We follow \cite{text2motion} 
to use this representation for both datasets. (2) KIT Motion-Language \cite{KITML} is a text-to-motion dataset that contains 3,911 human motion sequences and
6,353 sentences of textual descriptions. Our dataset split protocol follows previous works \cite{mld, mdm, temos}.

\noindent \textbf{Evaluation Metrics.} Following \cite{mdm,mld,temos,text2motion}, we
use five standard metrics. (1) Frechet Inception Distance (FID), which is the similarity between the generated motion features and the ground truth; (2) R-Precision, designed to assess the congruence between motions and text features. It quantifies this alignment by evaluating the retrieval precision of the top 3 pairs with the closest Euclidean distances between motion and text features; (3) Diversity, which calculates the mean Euclidean distance of random motion pairs; (4) MultiModality Distance (MMD), which is the average Euclidean distance between features of each generated motion and the corresponding text; and (5) MultiModality, which is the diversity
between each generated motion under the same text description.

\vspace{1mm}
\noindent \textbf{Baselines}. We compare our results to the following baselines:
TEMOS \cite{temos}, T2M \cite{text2motion}, TM2T \cite{tm2t}, MDM \cite{mdm}, MLD \cite{mld}, MotionDiffuse \cite{motiondiffuse}, DiffKFC \cite{diffkfc}, T2MGPT \cite{t2mgpt}, Fg-T2M \cite{fgt2m}, MotionGPT \cite{motiongpt}, M2DM \cite{M2DM}, and AttT2M \cite{attt2m}. Among them, the MLD is the baseline model for latent diffusion comparison and the DiffKFC is the baseline that uses the keyframes method. The AttT2M serves as the SOTA baseline.

\vspace{-2mm}
\begin{table*}[htp]
    \renewcommand{\arraystretch}{1}
    \centering
    \caption{\textbf{HumanML3D dataset.} Metrics are suggested by Guo et al. \cite{text2motion}. To reduce the influence of random error, we repeat the evaluation 20 times and present the mean statistic with a 95\% confidence interval, assuming the result follows normal distribution. We mark the best results with \textbf{bold} and the second best with \underline{underlines}. For DiffKFC, we present their results with no ground truth keyframes leaked to align with our method. MDO is the Maximum Distance Optimal keyframe and UNI is the uniform keyframe.}
    \resizebox{1\textwidth}{!}{%
    \begin{small}
    \begin{tabular}{lccccccc}
    \hline
    \hline
    \multirow{2}{*}{Models} & \multicolumn{3}{c}{R-precision $\uparrow$} & \multirow{2}{*}{FID$\downarrow$} & \multirow{2}{*}{MultiModal Dist$\downarrow$} & \multirow{2}{*}{Diversity$\uparrow$} & \multirow{2}{*}{MultiModality$\uparrow$} \\
    \cline{2-4}
    & Top-1& Top-2& Top-3 & & & & \\ 
    \hline
    \textbf{Real} & $0.511^{\pm .003}$ & $0.703^{\pm .003}$ & $0.797^{\pm .002}$& $0.002^{\pm .000}$& $2.974^{\pm .008}$& $9.503^{\pm .065}$& -\\
    \hline
    TEMOS\cite{temos} &$0.424^{\pm .002}$  &  $0.612^{\pm .002}$ & $0.722^{\pm .002}$ & $3.734^{\pm .028}$ &$3.703^{\pm .008}$ & $8.973^{\pm .071}$&$0.368^{\pm .018}$\\
    TM2T\cite{tm2t} & $0.424^{\pm .003}$  & $0.618^{\pm .003}$   &$0.729^{\pm .002}$  &$1.501^{\pm .017}$& $3.467^{\pm .011}$ & $8.589^{\pm .076}$ & $2.424^{\pm .093}$\\
    T2M\cite{text2motion} & $0.455^{\pm .003}$  & $0.636^{\pm .003}$   &$0.736^{\pm .002}$  &$1.087^{\pm .021}$& $3.347^{\pm .008}$ & $9.175^{\pm .083}$ & $2.219^{\pm .074}$\\
    MDM\cite{mdm} & $0.320^{\pm .005}$ &  $0.498^{\pm .004}$ &  $0.611^{\pm .007}$ &  $0.544^{\pm .044}$ &  $5.566^{\pm .027}$ & {$9.559^{\pm .086}$} &   {$2.799^{\pm .072}$}\\
    MLD\cite{mld} & $0.481^{\pm .003}$ & $0.673^{\pm .003}$ &  $0.772^{\pm .002}$ &  $0.473^{\pm .013}$ &  $3.196^{\pm .010}$ & $9.724^{\pm .082}$ &   $2.413^{\pm .079}$\\
    MotionDiffuse\cite{motiondiffuse}& {$0.491^{\pm .001}$} & $0.681^{\pm .001}$ &  {$0.782^{\pm .001}$} &  $0.630^{\pm .001}$ &  $3.113^{\pm .001}$ & $9.410^{\pm .049}$ &   $1.553^{\pm .042}$\\
    DiffKFC\cite{diffkfc} + 0\% & - & - & $0.602^{\pm .008}$ & $0.597^{\pm .025}$ & $5.721^{\pm .036}$ & {$9.576^{\pm .049}$} & $\underline{2.984}^{\pm .054}$\\
    T2MGPT\cite{t2mgpt} & {$0.491^{\pm .003}$} & $0.680^{\pm .003}$  & $0.775^{\pm .002}$ &  $\underline{0.116}^{\pm .004}$& $3.118^{\pm .011}$ & {$9.761^{\pm .081}$} &$1.856^{\pm .011}$\\
    Fg-T2M\cite{fgt2m}&  {$0.492^{\pm .002}$} & {$0.683^{\pm .003}$} & {$0.783^{\pm .002}$} & {$0.243^{\pm .019}$} & {$3.109^{\pm .007}$} & $9.278^{\pm .072}$ & $1.619^{\pm .049}$\\
    MotionGPT\cite{motiongpt}&  {$0.492^{\pm .003}$} & {$0.681^{\pm .003}$} & {$0.778^{\pm .003}$} & {$0.232^{\pm .008}$} & {$3.096^{\pm .012}$} & $9.528^{\pm .052}$ & {$2.008^{\pm .075}$}\\
    M2DM\cite{M2DM}&  $\underline{0.497}^{\pm .003}$ & {$0.682^{\pm .002}$} & {$0.763^{\pm .003}$} & {$0.352^{\pm .005}$} & {$3.134^{\pm .010}$} &$ \mathbf{9.926}^{\pm .073}$ & $\mathbf{3.587}^{\pm .072}$\\
    AttT2M\cite{attt2m}&  $\mathbf{0.499}^{\pm .003}$ & $\underline{0.690}^{\pm .002}$ & $\underline{0.786}^{\pm .002}$ & $\textbf{0.112}^{\pm .006}$ & $\underline{3.038}^{\pm .007}$ & {$9.700^{\pm .090}$} & {$2.452^{\pm .051}$}\\
    \hline
    Ours, skip-only (MDO) &  $0.493^{\pm .004}$ & {$0.689^{\pm .003}$} & $0.784^{\pm .003}$ & $0.199^{\pm .008}$ & {$3.088^{\pm .020}$} & {$9.742^{\pm .050}$} & $2.025^{\pm .073}$\\
    Ours, parallel-only (MDO) &  $0.495^{\pm .003}$ & $0.682^{\pm .003}$ & $0.782^{\pm .003}$ & $0.216^{\pm .012}$ & $3.063^{\pm .019}$ & $9.714^{\pm .064}$ & $1.974^{\pm .080}$\\
    Ours, parallel-skip (UNI) &  {$0.494^{\pm .003}$} & {$0.688^{\pm .003}$} & $\underline{0.786}^{\pm .003}$ & {$0.178^{\pm .008}$} & {$3.087^{\pm .012}$} & $9.594^{\pm .052}$ & $1.988^{\pm .075}$\\
    Ours, parallel-skip (MDO) &  $\textbf{0.499}^{\pm .003}$ & $\mathbf{0.700}^{\pm .003}$ & $\mathbf{0.794}^{\pm .003}$ & {$0.192^{\pm .008}$} & $\mathbf{3.036}^{\pm .010}$ & $\underline{9.803}^{\pm .100}$ & $2.670^{\pm .063}$\\

    \hline
    \end{tabular}
    \end{small}
    }
    \vspace{-5mm}
    \label{tab:humanml3d}
\end{table*}

\subsection{Implementation Details}\label{subsec:implement}
\noindent \textbf{Keyframe VAE}  
has 8 layers of vanilla Transformer Encoder and 8 layers of Transformer Decoder. Each Transformer layer has 272 dimensions (256 for keyframe embedding and 16 for keyframe index embedding), 1024 hidden dimensions, 4 heads, a drop-out rate of 10\%, and ReLU activation. This module is trained for 4000 epochs. The VAE is trained using a learning rate of $5\times10^{-5}$ in the first 2000 epochs and $2\times 10^{-5}$ in the last 2000 epochs. 
%
\textbf{PST Denoiser} has $N_s=2$ Transformer encoder layers and $N_c=5$ cross-attention layers. Each Transformer encoder layer follows the same configuration as the VAE. For the ablation study, the skip-only Transformer has 11 layers, to match it as closely as possible to the full PST. Chen et al. \cite{mld} show that the number of layers only slightly influences the results. Our PST employs the CLIP-VIT-LARGE-14 model with frozen weights as the text encoder. Training is done for 1000 epochs. 
%
\textbf{MMAE} is an 8-layer text-conditioned residual Transformer. The keyframe ratio is set to 10\% and 12.5\% in uniform intervals for HumanML3D \cite{text2motion} and KIT \cite{KITML} datasets, respectively. To fully learn the redundant pose representation, we set the dimension of the Transformer layer to 512, the number of heads to 8, the hidden dimension to 2048, and used a 0.1 drop-out ratio. The text condition is concatenated with the masked motion sequence before passing to the MMAE which is trained for 1000 epochs with a constant learning rate of $5\times10^{-5}$.

\subsection{Text-to-Motion Generation Results}\label{subsec:compt2m}
Table \ref{tab:humanml3d} compares our results to existing methods on the HumanML3D dataset. The metrics are arranged according to their importance; the left column being the most important. {The first row (Real) presents the metrics calculated for the ground truth motions on the test dataset for reference.}
We can see that our model achieves the best results on all three R-precision metrics, as well as multi-modal distance. Our method also achieves competitive performance on the FID metric. 
These results show that our method generates human motions that are better aligned with the input text.

\begin{table}[h!]
    \renewcommand{\arraystretch}{1.25}
    \centering
    \caption{KeyMotion performance on the \textbf{HumanML3D} with different keyframe rates. The latent dimension for all keyframe rates is 2 using PST. When the keyframe ratio is 100\%, no MMAE is employed for motion-filling.}
    \resizebox{0.48\textwidth}{!}{%
    \begin{footnotesize}
    \begin{tabular}{lccccc}
    \hline
    \hline
    \multirow{2}{*}{Keyframe Rate} & \multicolumn{3}{c}{R-precision $\uparrow$} & \multirow{2}{*}{FID$\downarrow$} & \multirow{2}{*}{MultiModal Dist$\downarrow$} \\
    \cline{2-4}
     & Top-1& Top-2& Top-3 & &  \\ 
    \hline
    \textbf{Real} & $0.511^{\pm .003}$ & $0.703^{\pm .003}$ & $0.797^{\pm .002}$& $0.002^{\pm .000}$& $2.974^{\pm .008}$\\

    5.00 \% &  $0.486^{\pm.003}$ & $0.679^{\pm.004}$ & $0.781^{\pm.003}$ & $0.364^{\pm.009}$ & $3.163^{\pm.008}$\\
    
    6.25 \% & $0.492^{\pm.004}$ & $0.682^{\pm.004}$ & $0.785^{\pm.003}$ & $0.283^{\pm.010}$ & $3.117^{\pm.009}$ \\
    
    8.33 \% & $0.495^{\pm.003}$ & $0.699^{\pm.004}$ & $0.789^{\pm.003}$ & $0.242^{\pm.010}$ & $3.084^{\pm.010}$\\
    
    10.0 \% &  $\mathbf{0.499^{\pm .003}}$ & $\mathbf{0.700^{\pm .003}}$ & $\mathbf{0.794^{\pm .003}}$ & $\mathbf{0.192^{\pm .008}}$ & $\mathbf{3.036^{\pm .010}}$\\

    12.5 \% &  $0.496^{\pm0.004}$ & $0.692^{\pm .004}$ & {$0.790^{\pm .003}$} & {$0.200^{\pm .009}$} & {$3.061^{\pm .009}$}\\

    25.0 \% &  $0.482^{\pm0.003}$ & $0.682^{\pm .003}$ & {$0.784^{\pm .003}$} & {$0.335^{\pm .013}$} & {$3.104^{\pm .011}$}\\

    50.0 \% &  $0.468^{\pm0.003}$ & $0.656^{\pm .003}$ & {$0.763^{\pm .003}$} & {$0.418^{\pm .012}$} & {$3.219^{\pm .012}$}\\
    100. \% &  $0.450^{\pm0.003}$ & $0.642^{\pm .003}$ & {$0.747^{\pm .003}$} & {$0.505^{\pm .008}$} & {$3.382^{\pm .011}$}\\
    \hline
    \end{tabular}
    \end{footnotesize}
    }
    \label{tab:kf}
    \vspace{-5mm}
\end{table}

We also investigated how the ratio of selected keyframes affects performance. Our results in Table \ref{tab:kf} show that 10\% keyframes achieves the best performance on all metrics. Interestingly, using all frames (100\%) reduces performance, which we believe is due to the close proximity problem we discussed in the Introduction.

Table \ref{tab:kit} shows results for the KIT dataset. Although our model does not achieve SOTA performance for Top-1 and Top-2 R-precision, it achieves the best performance for Top-3 R-precision. Interestingly, the parallel-only variant of our method achieves the highest Diversity and the second-best FID metric. 
Looking at both tables, we see that our method and its variants are more consistent in achieving the best and second-best performances on most metrics, which shows its better generalization ability compared to the baselines.
A possible reason for lower results on the KIT dataset is that it has small motion lengths and fewer training samples which limits the infilling ability of our MMAE; HumanML3D/KIT have 140/73 average frames per sequence and 24,546/4,394 training sequences (after data augmentation). 

\begin{table*}[t!]
    \renewcommand{\arraystretch}{1}
    \centering
    \caption{\textbf{KIT dataset.} Metrics are suggested by Guo et al. \cite{text2motion}. To mitigate the impact of random errors, evaluation is performed 20 times and the mean statistics are presented with a 95\% confidence interval, assuming that the results follow normal distributions. The best results are highlighted in \textbf{bold}, and the second-best with \underline{underlines}.
    MDO is the Maximum Distance Optimal keyframe and UNI is the uniform keyframe.}
    \resizebox{1\textwidth}{!}{%
    \begin{small}
    \begin{tabular}{lccccccc}
    \hline
    \hline
    \multirow{2}{*}{Models} & \multicolumn{3}{c}{R-precision $\uparrow$} & \multirow{2}{*}{FID$\downarrow$} & \multirow{2}{*}{MultiModal Dist$\downarrow$} & \multirow{2}{*}{Diversity$\uparrow$} & \multirow{2}{*}{MultiModality$\uparrow$} \\
    \cline{2-4}
    & Top-1& Top-2& Top-3 & & & & \\ 
    \hline
    \textbf{Real} & $0.424^{\pm .005}$ & $0.649^{\pm .006}$ & $0.779^{\pm .006}$& $0.031^{\pm .004}$& $2.788^{\pm .012}$& $11.08^{\pm .097}$& -\\
    \hline
    TEMOS\cite{temos} &$0.353^{\pm .002}$  &  $0.561^{\pm .002}$ & $0.687^{\pm .002}$ & $3.717^{\pm .028}$ &$3.417^{\pm .008}$ & $10.84^{\pm .071}$&$0.532^{\pm .018}$\\
    TM2T\cite{tm2t} & $0.280^{\pm .006}$  & $0.463^{\pm .007}$   &$0.587^{\pm .005}$  &$3.599^{\pm .051}$& $4.591^{\pm .019}$ & $9.473^{\pm .100}$ & $\underline{3.292}^{\pm .034}$\\
    T2M\cite{text2motion} & $0.361^{\pm .006}$  & $0.559^{\pm .007}$   &$0.681^{\pm .007}$  &$3.022^{\pm .107}$& $3.488^{\pm .028}$ & $10.72^{\pm .145}$ & $2.052^{\pm .107}$\\
    MDM\cite{mdm} &   $0.164^{\pm .004}$ &   $0.291^{\pm .004}$ &  $0.396^{\pm .004}$ &  $0.497^{\pm .021}$ &  $9.191^{\pm .022}$ & {$10.85^{\pm .107}$} &   {$1.907^{\pm .214}$}\\
    MLD\cite{mld} & $0.390^{\pm .008}$ & $0.609^{\pm .008}$ &  $0.734^{\pm .007}$ &  {$0.404^{\pm .037}$} &  $3.204^{\pm .027}$ & $10.80^{\pm .117}$ &   {$2.192^{\pm .074}$}\\
    DiffKFC\cite{diffkfc} & - & - & $0.420^{\pm.007}$& $\mathbf{0.164}^{\pm.026}$ & - & $10.98^{\pm.108}$ & -     \\
    MotionDiffuse\cite{motiondiffuse}& $\underline{0.417}^{\pm .004}$ & {$0.621^{\pm .004}$} &  {$0.739^{\pm .004}$} &  $1.954^{\pm .062}$ &  $\mathbf{2.958}^{\pm .005}$ & $11.10^{\pm .143}$ &   $0.730^{\pm .013}$\\
    T2MGPT\cite{t2mgpt} & {$0.416^{\pm .006}$} & {$0.627^{\pm .006}$}  & {$0.745^{\pm .006}$} &  $0.514^{\pm .029}$& $\underline{3.007}^{\pm .023}$ & $10.92^{\pm .108}$ &$1.570^{\pm .039}$\\
    Fg-T2M\cite{fgt2m} & $\mathbf{0.418}^{\pm .005}$ & {$0.626^{\pm .004}$}  & {$0.745^{\pm .004}$} &  $0.571^{\pm .047}$& {$3.114^{\pm .015}$} & $10.93^{\pm .083}$ &$1.856^{\pm .011}$\\
    MotionGPT\cite{motiongpt} & {$0.366^{\pm .005}$} & {$0.510^{\pm .004}$}  & {$0.680^{\pm .005}$} &  $0.571^{\pm .047}$& {$3.527^{\pm .021}$} & $10.35^{\pm .084}$ &{$2.328^{\pm .117}$}\\
    M2DM\cite{M2DM} & {$0.416^{\pm .004}$} & $\underline{0.628}^{\pm .004}$  & {$0.743^{\pm .004}$} &  $0.515^{\pm .029}$& {$3.015^{\pm .017}$} & $\underline{11.42}^{\pm .097}$ & $\mathbf{3.325}^{\pm .037}$\\
    AttT2M\cite{motiongpt} & {$0.413^{\pm .005}$} & $\mathbf{0.632}^{\pm .004}$  & $\underline{0.751}^{\pm .005}$ &  $0.870^{\pm .039}$& {$3.039^{\pm .021}$} & $10.96^{\pm .123}$ &$2.281^{\pm .047}$\\
    \hline

    Ours, skip only (MDO)&  $0.399^{\pm .004}$ & $0.610^{\pm .005}$ & $0.730^{\pm .005}$ & {$0.474^{\pm .035}$} & $3.162^{\pm .036}$ & {$11.35^{\pm .099}$} & $2.125^{\pm .073}$\\

    Ours, parallel only (MDO) &  ${0.402^{\pm .005}}$ & $0.614^{\pm .005}$ & {$0.740^{\pm .004}$} & $\underline{0.386}^{\pm .030}$ & $3.124^{\pm .030}$ & $\mathbf{11.44}^{\pm .100}$ & $2.134^{\pm .070}$\\
    Ours, parallel-skip (UNI)&  $0.400^{\pm.004}$ & $0.613^{\pm .004}$ & {$0.749^{\pm .004}$} & {$0.572^{\pm .039}$} & {$3.028^{\pm .040}$} & {$11.30^{\pm .102}$} & $2.061^{\pm .076}$\\
    Ours, parallel-skip (MDO)&  $0.408^{\pm.004}$ & $0.621^{\pm .005}$ & $\mathbf{0.754^{\pm .004}}$ & {$0.566^{\pm .041}$} & {$3.014^{\pm .033}$} & {$10.89^{\pm .128}$} & $2.015^{\pm .066}$\\
    \hline

    \end{tabular}
    \end{small}
    }
    \vspace{-3mm}
    \label{tab:kit}
\end{table*}

\begin{figure*}[tp!]
  \centering
  \renewcommand{\arraystretch}{1.5}
  \resizebox{0.97\textwidth}{!}{%
  \begin{tabular}{ccccc}
    \hline
    \textbf{Motion (A)} & \textbf{Motion (B)} & \textbf{Motion (C)} & \textbf{Motion (D)}& \textbf{Motion (E)} \\
    \hline
    MLD\includegraphics[width=0.18\linewidth]{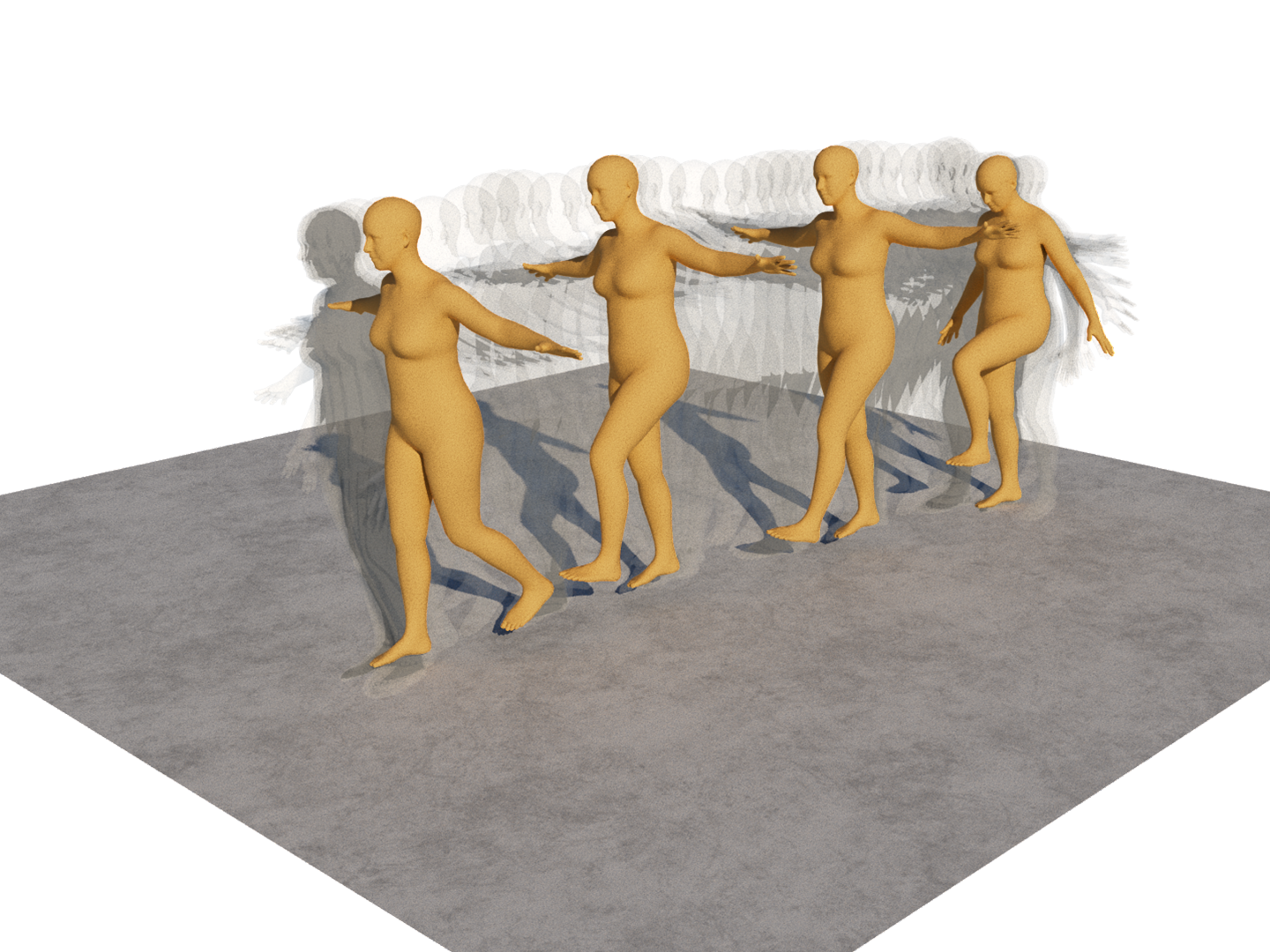} & \includegraphics[width=0.18\linewidth]{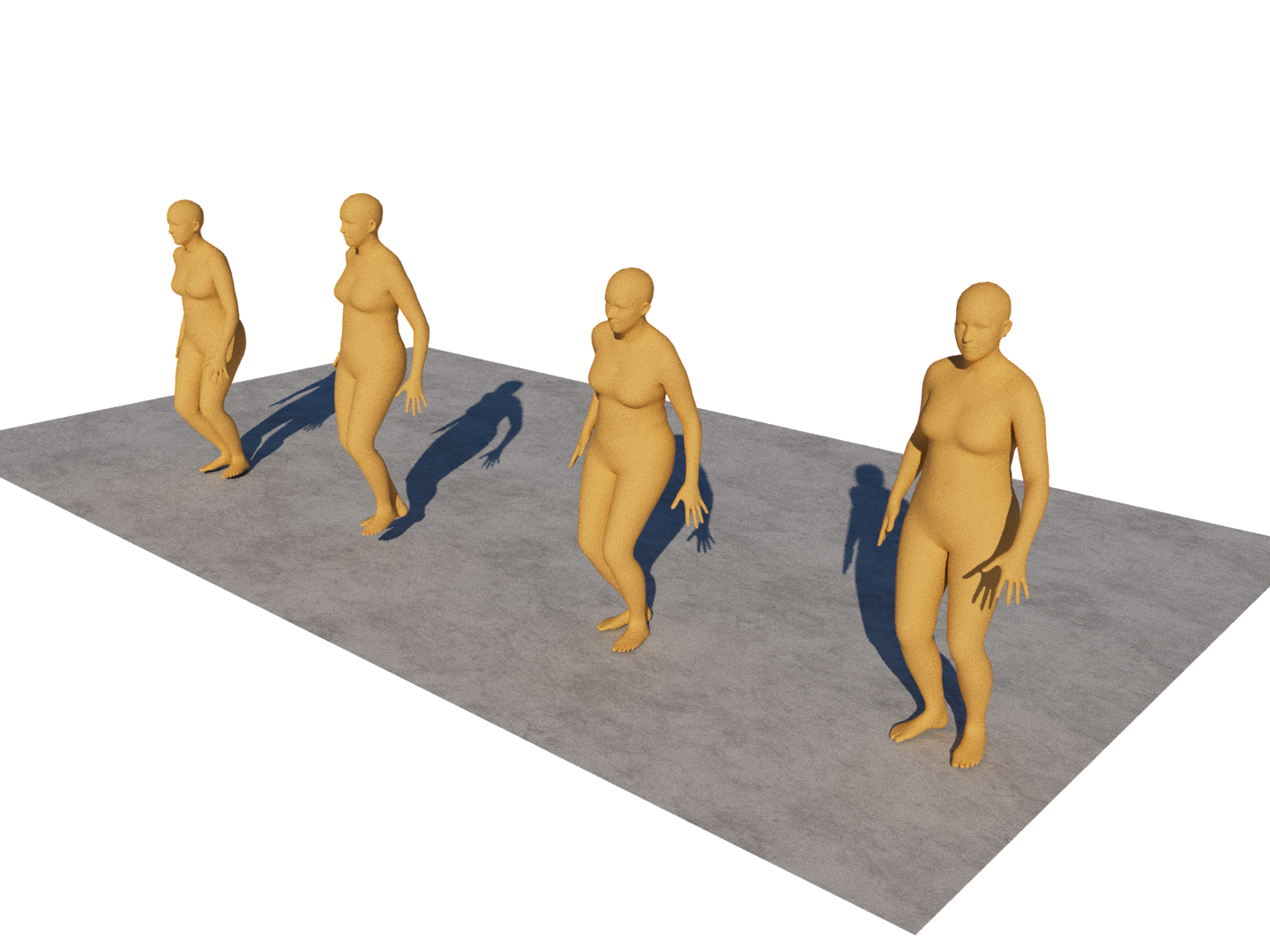} & \includegraphics[width=0.18\linewidth]{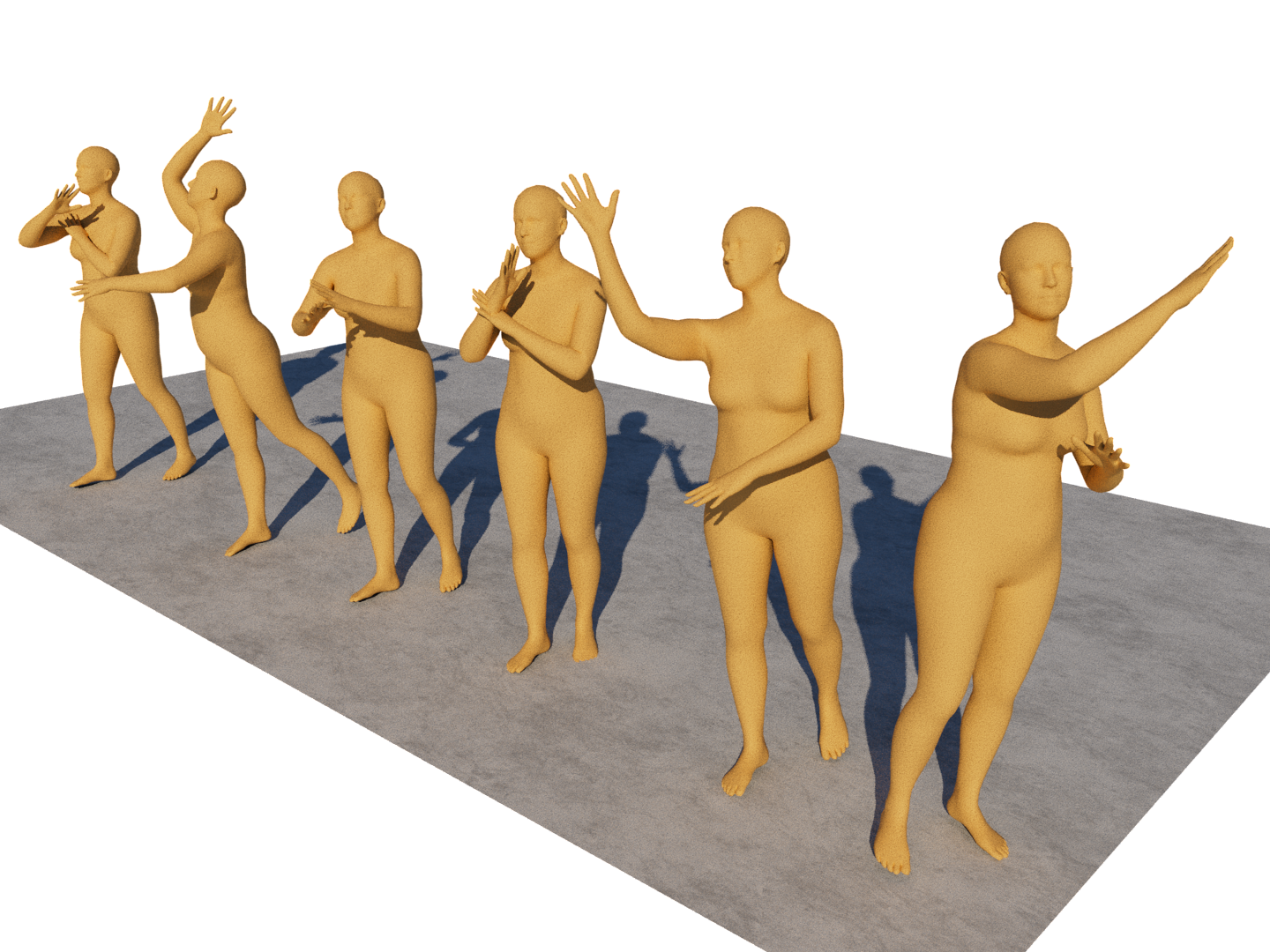} & \includegraphics[width=0.18\linewidth]{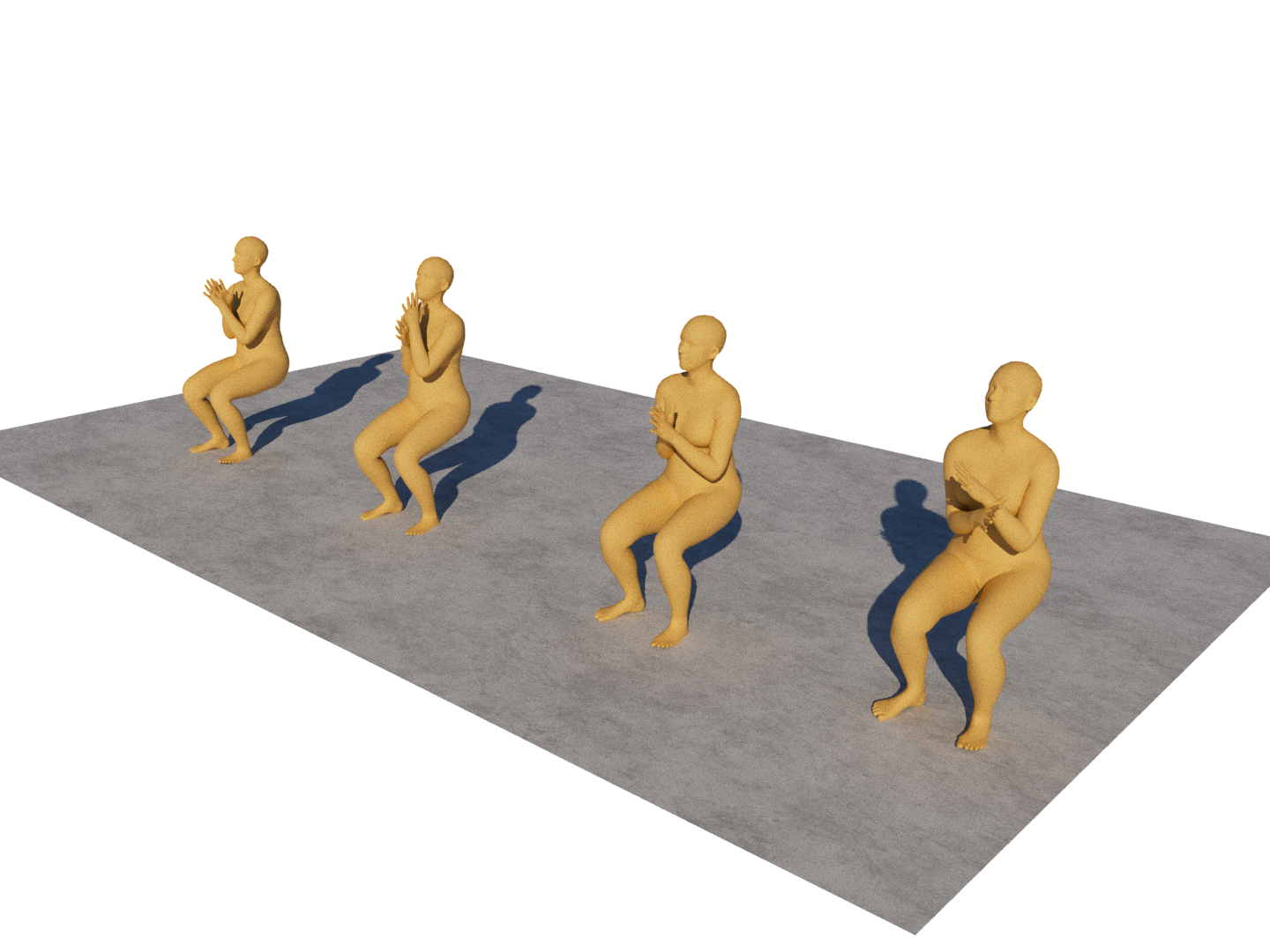} & \includegraphics[width=0.18\linewidth]{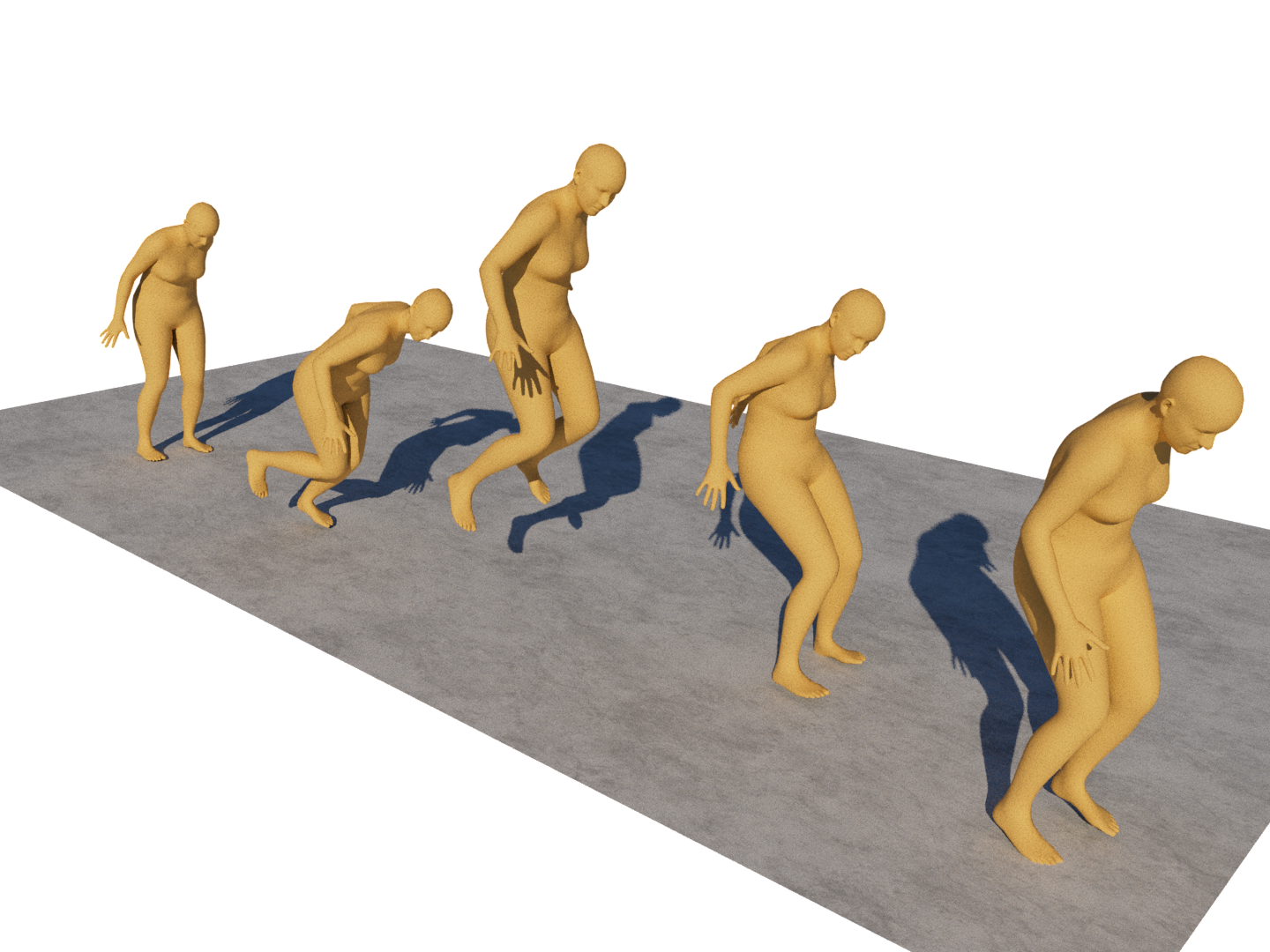}\\
    \hline
    T2MGPT\includegraphics[width=0.18\linewidth]{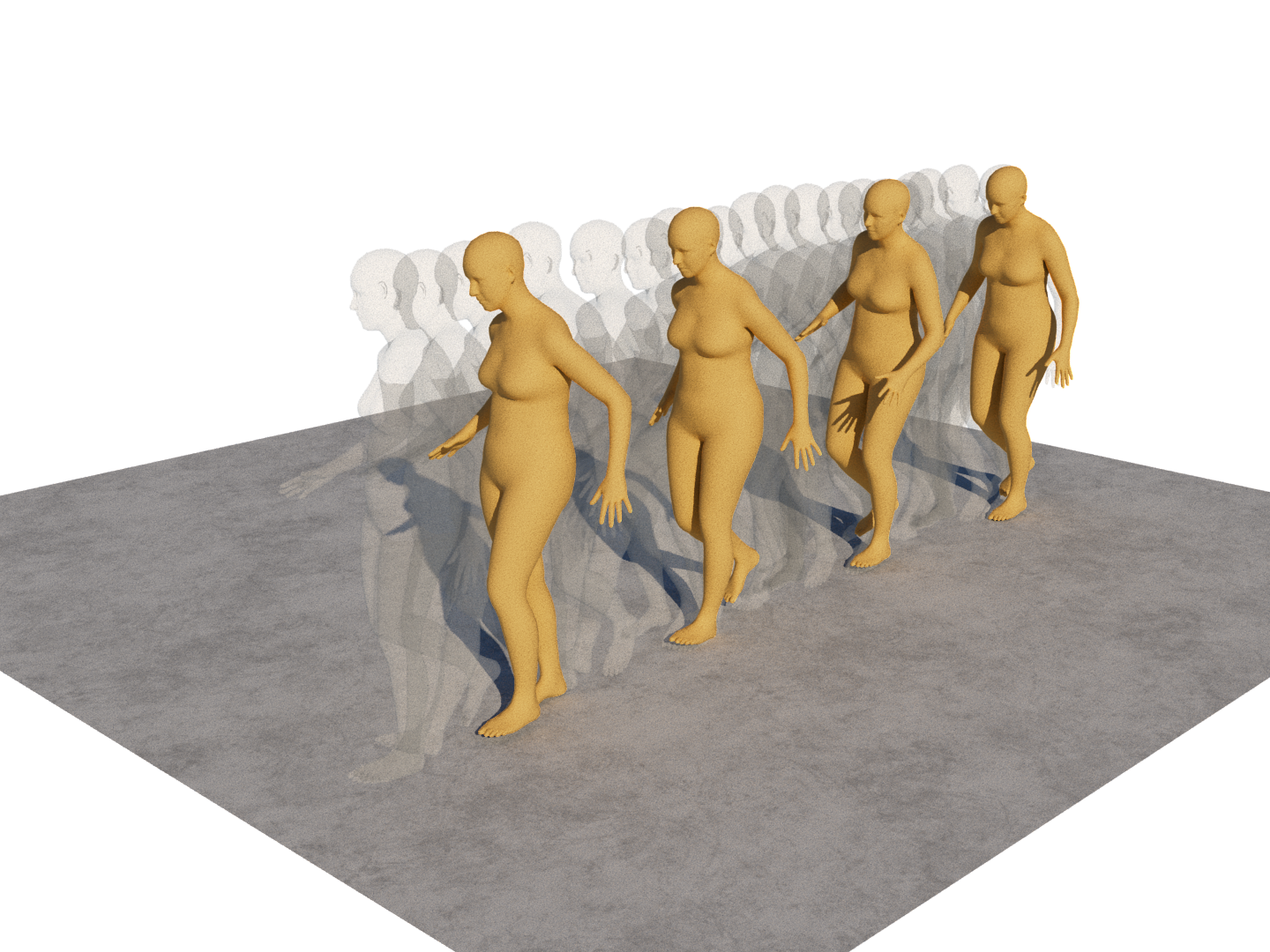} & \includegraphics[width=0.18\linewidth]{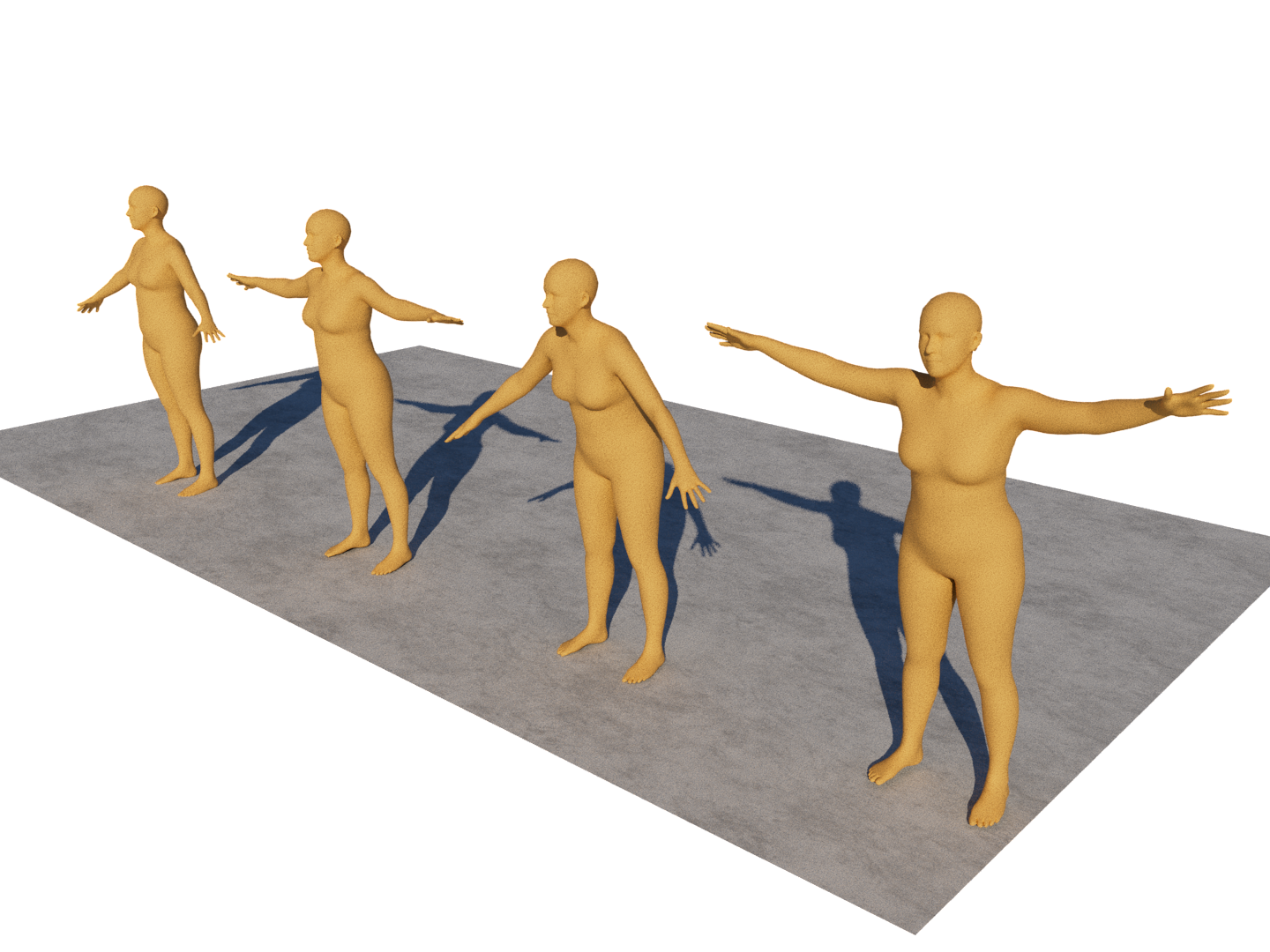} & \includegraphics[width=0.18\linewidth]{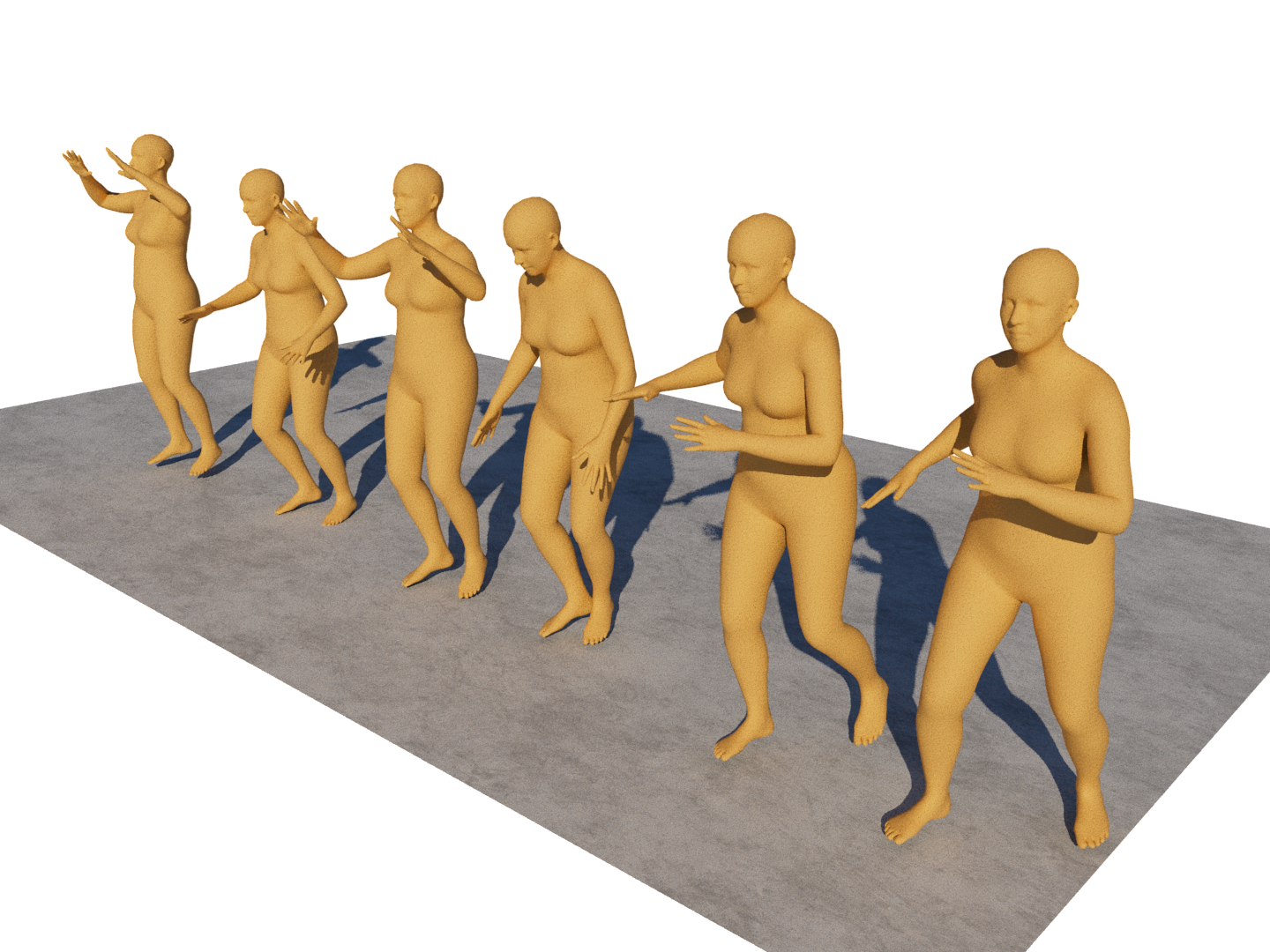} & \includegraphics[width=0.18\linewidth]{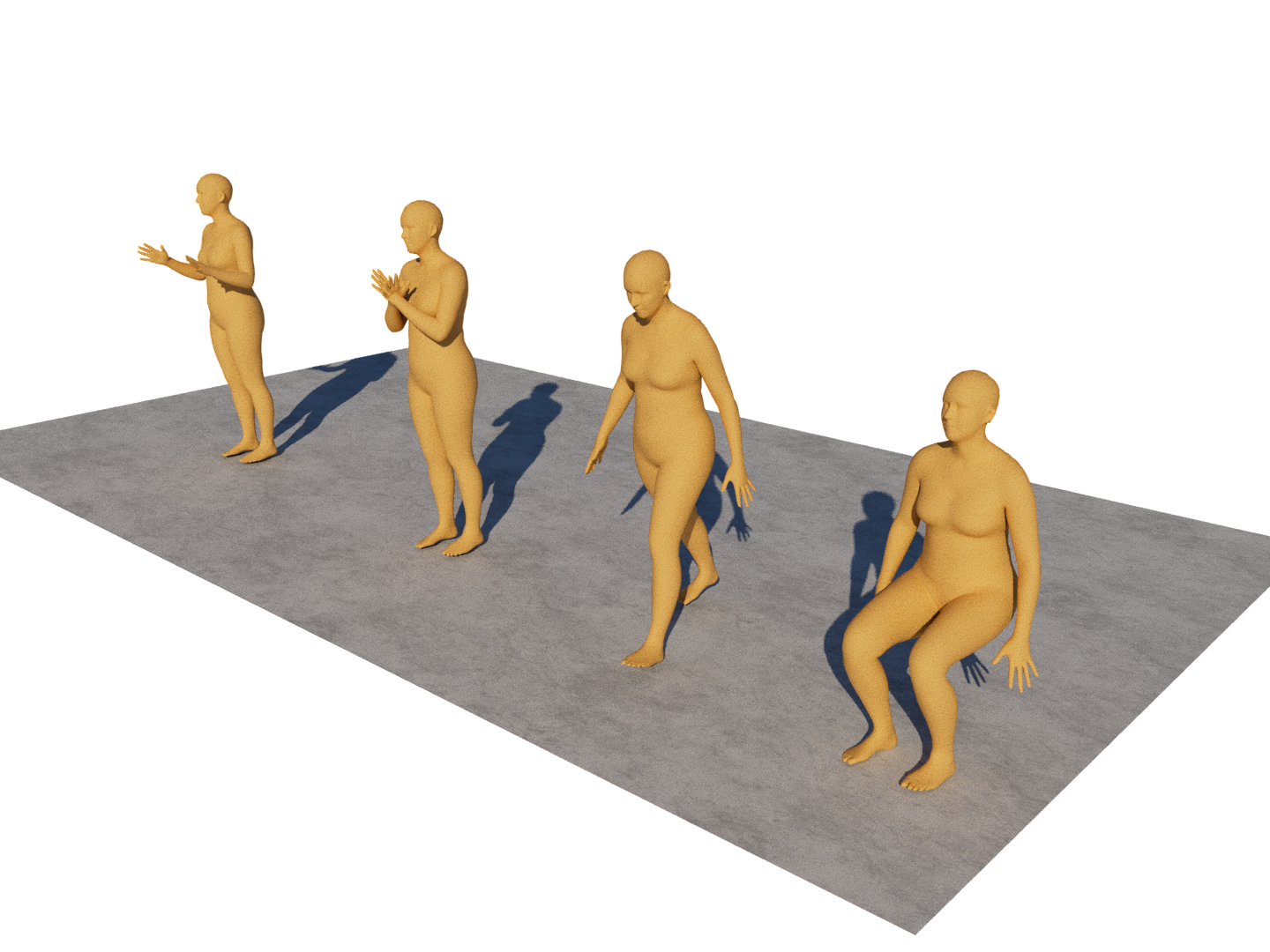} & \includegraphics[width=0.18\linewidth]{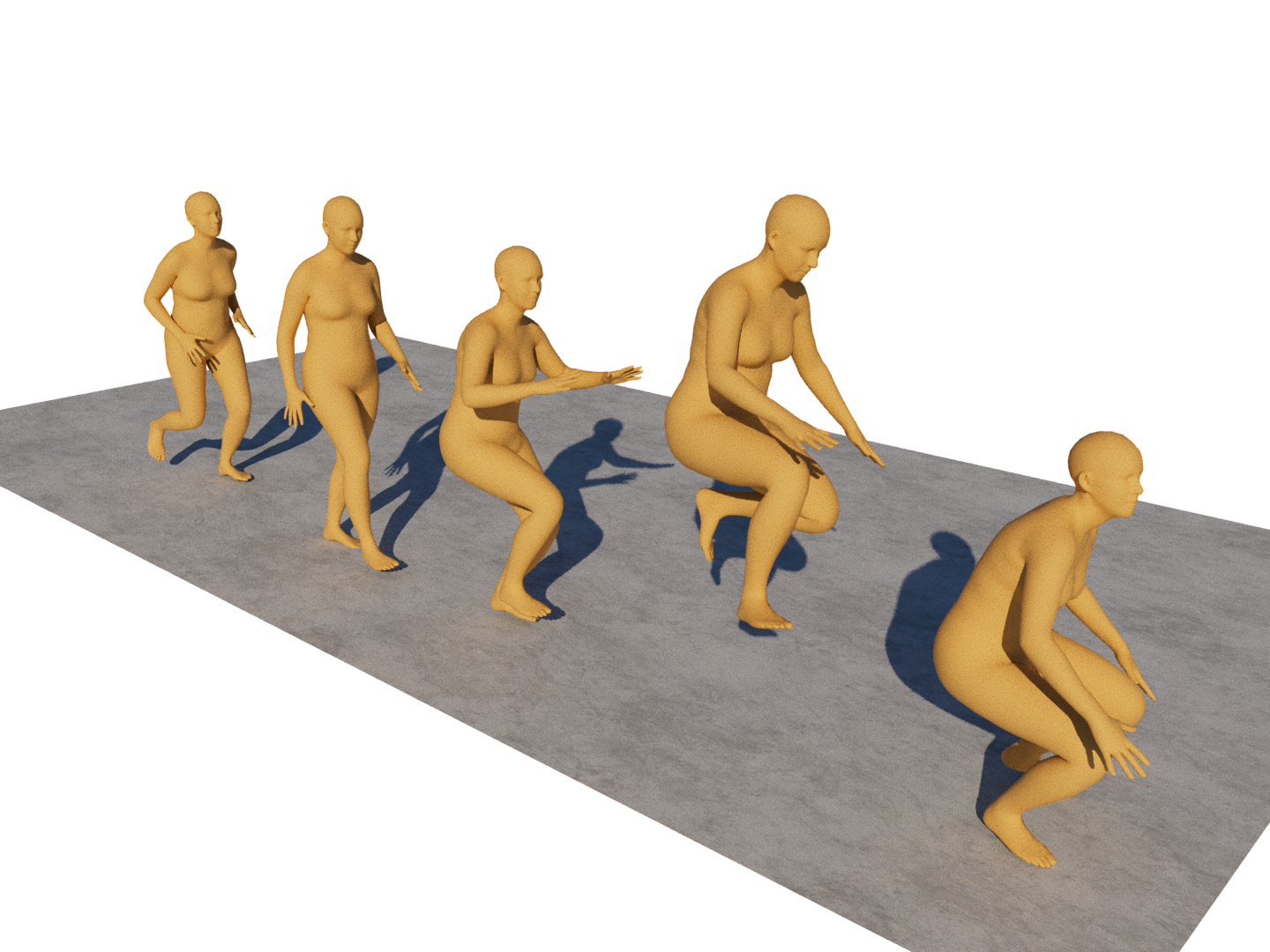}\\
    \hline
    AttT2M\includegraphics[width=0.18\linewidth]{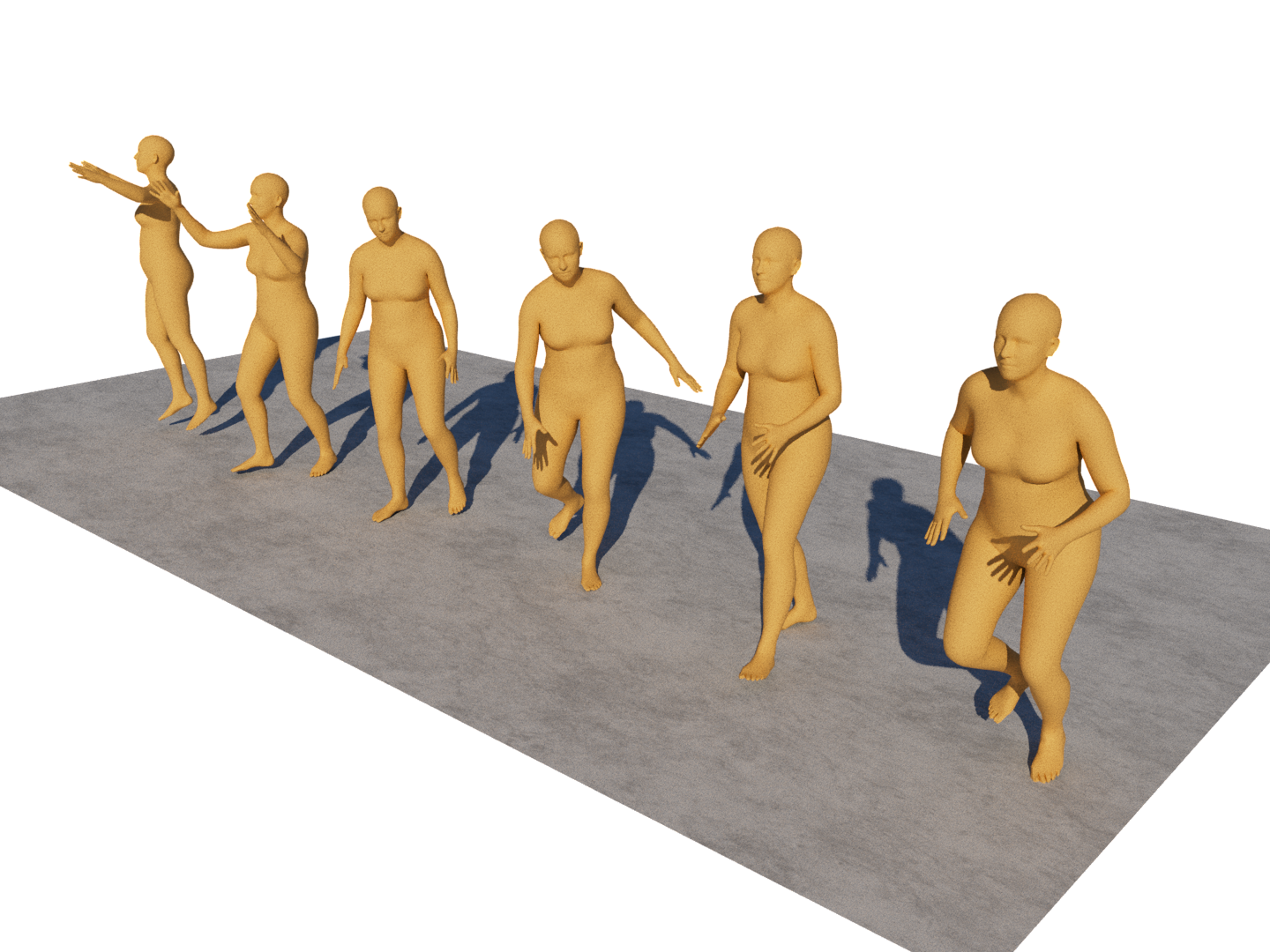} & \includegraphics[width=0.18\linewidth]{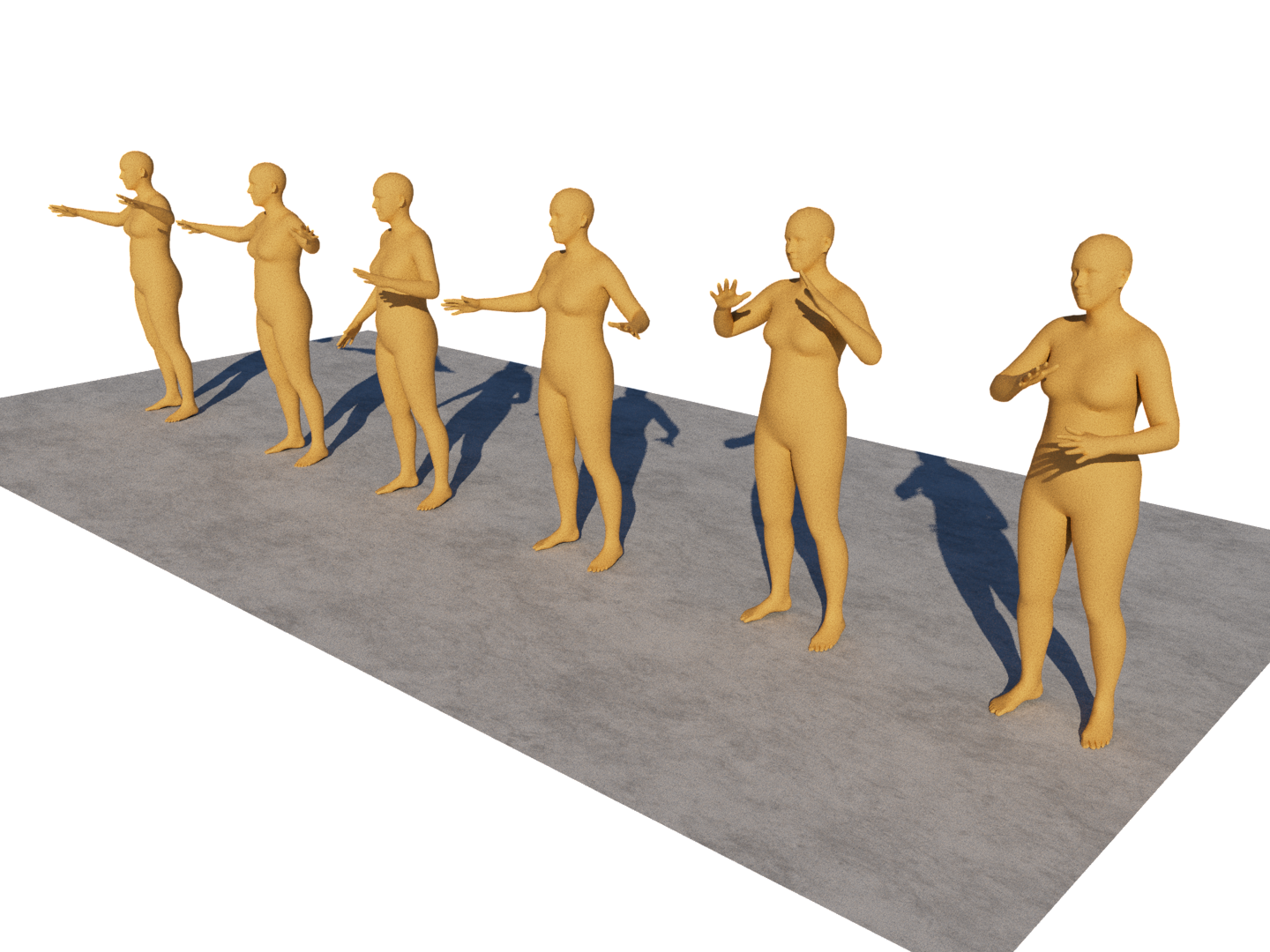} & \includegraphics[width=0.18\linewidth]{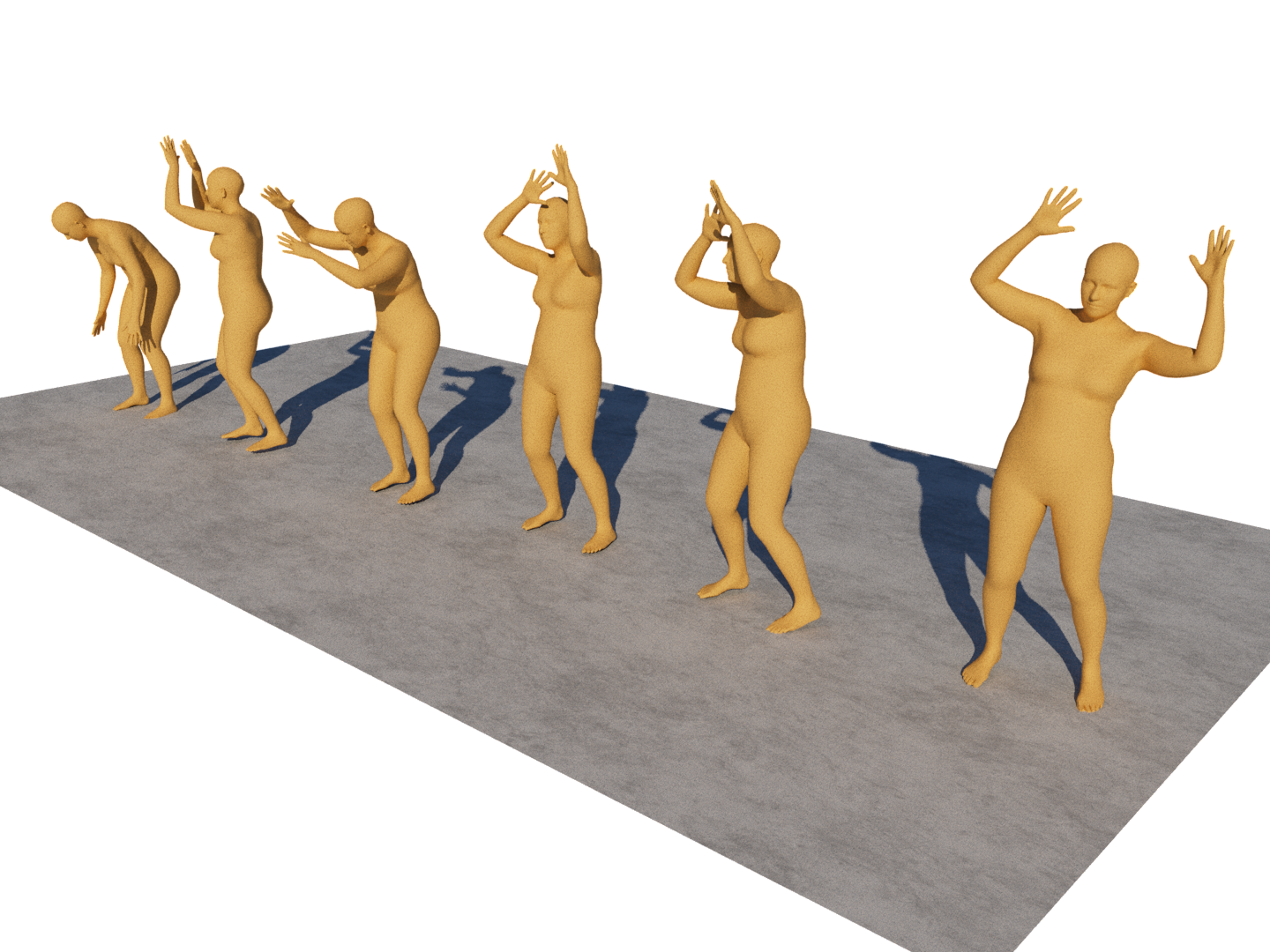} & \includegraphics[width=0.18\linewidth]{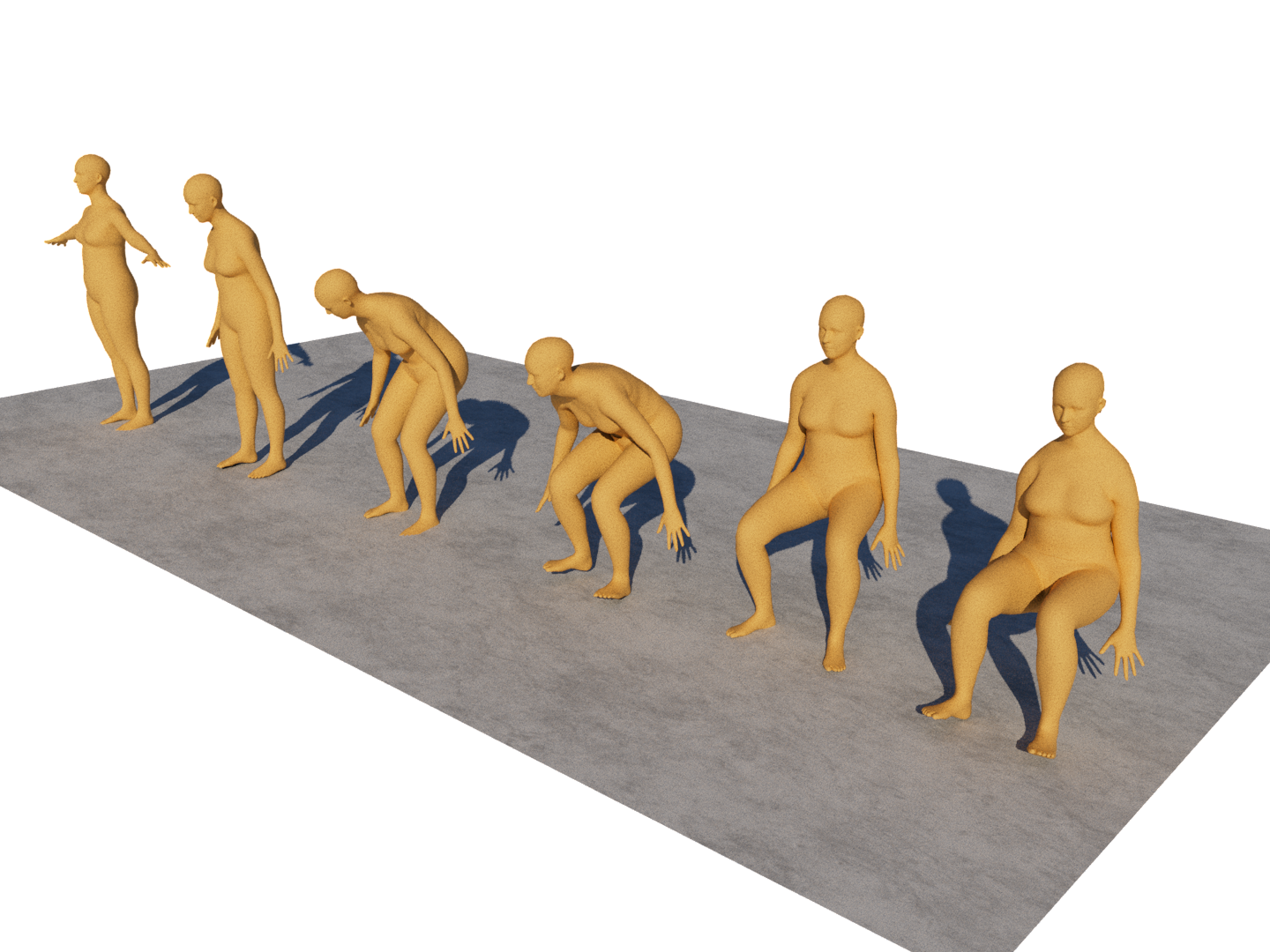} & \includegraphics[width=0.18\linewidth]{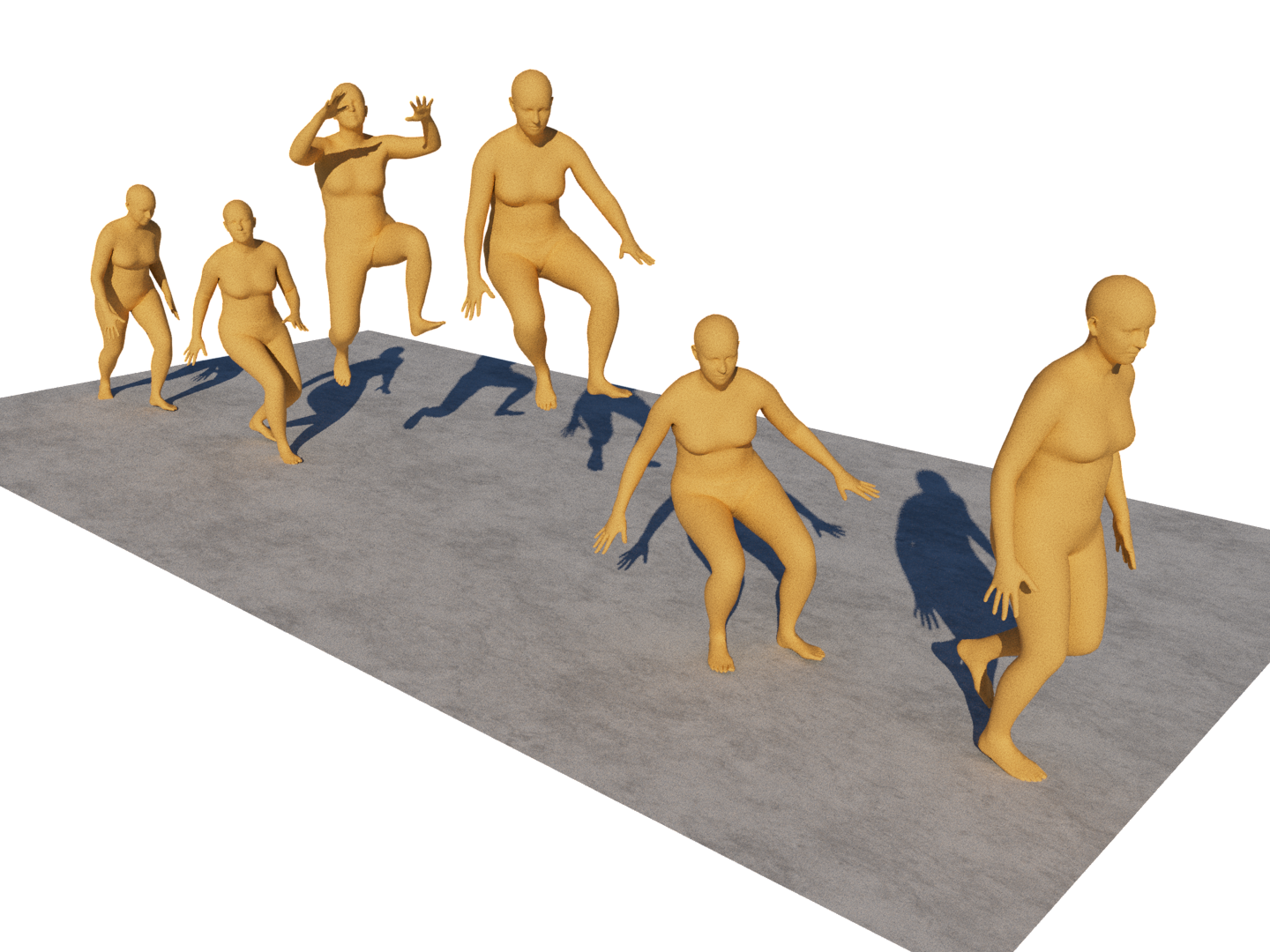}\\
    \hline
    Ours\includegraphics[width=0.18\linewidth]{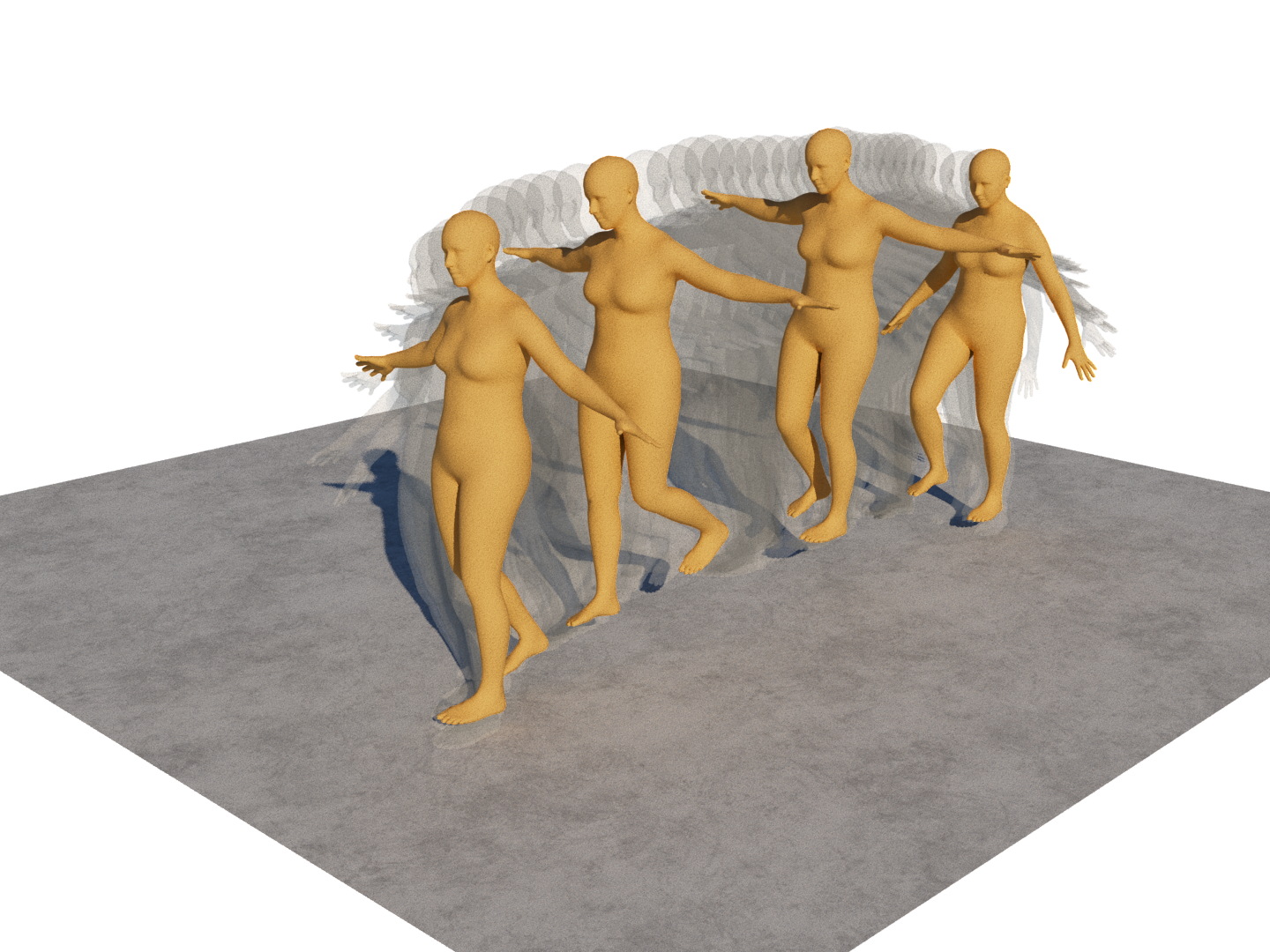} & \includegraphics[width=0.18\linewidth]{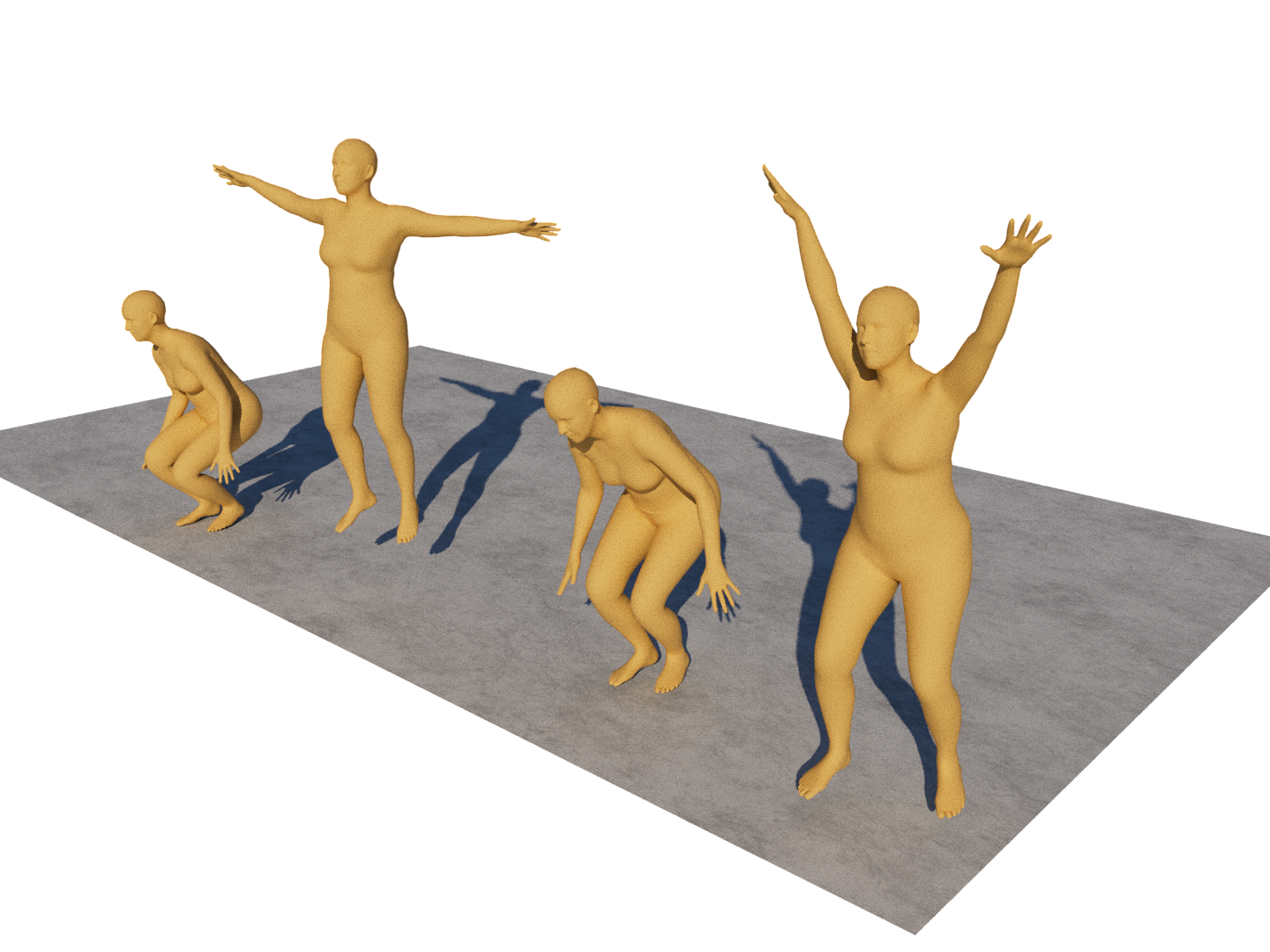} & \includegraphics[width=0.18\linewidth]{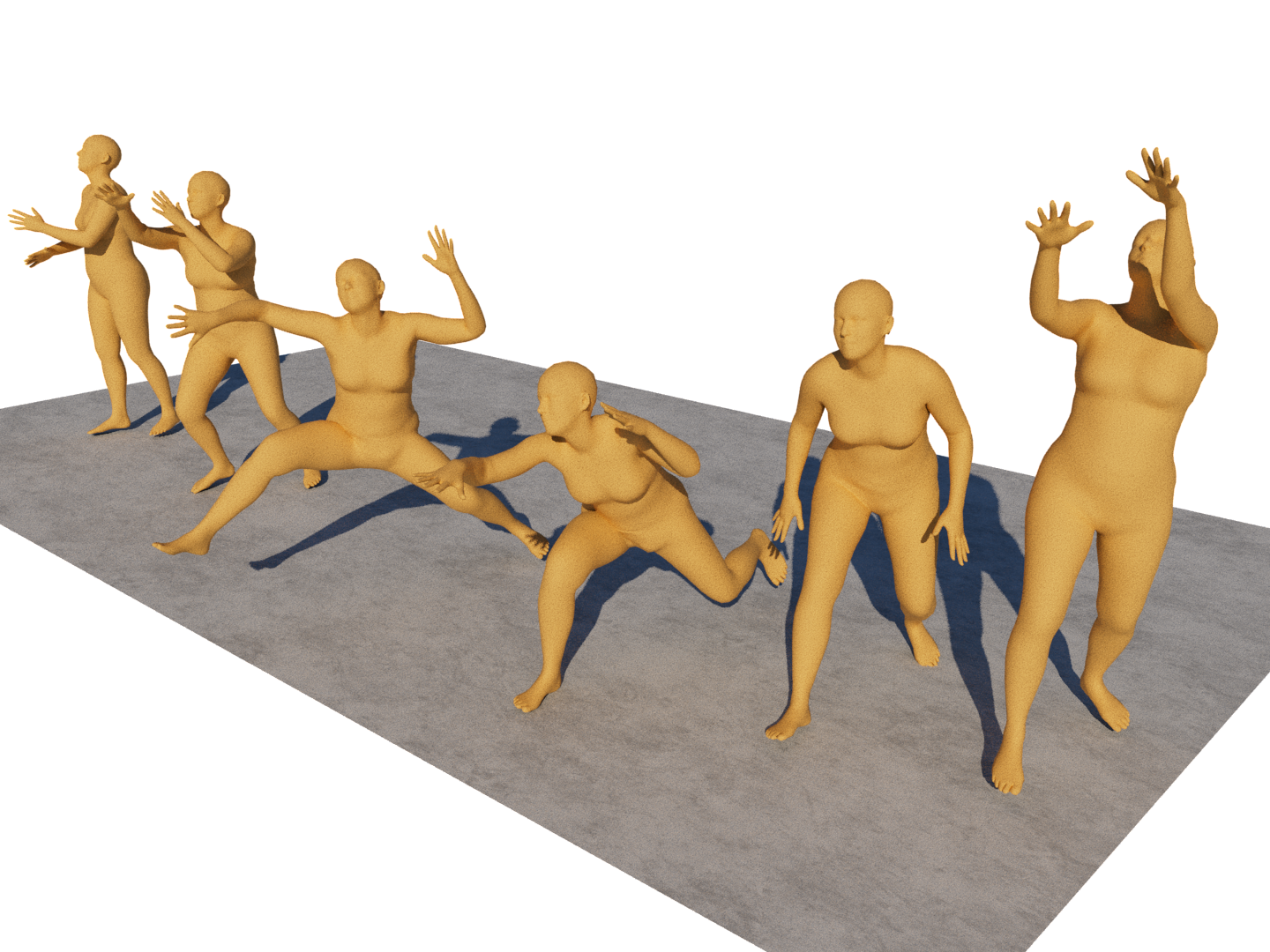} & \includegraphics[width=0.18\linewidth]{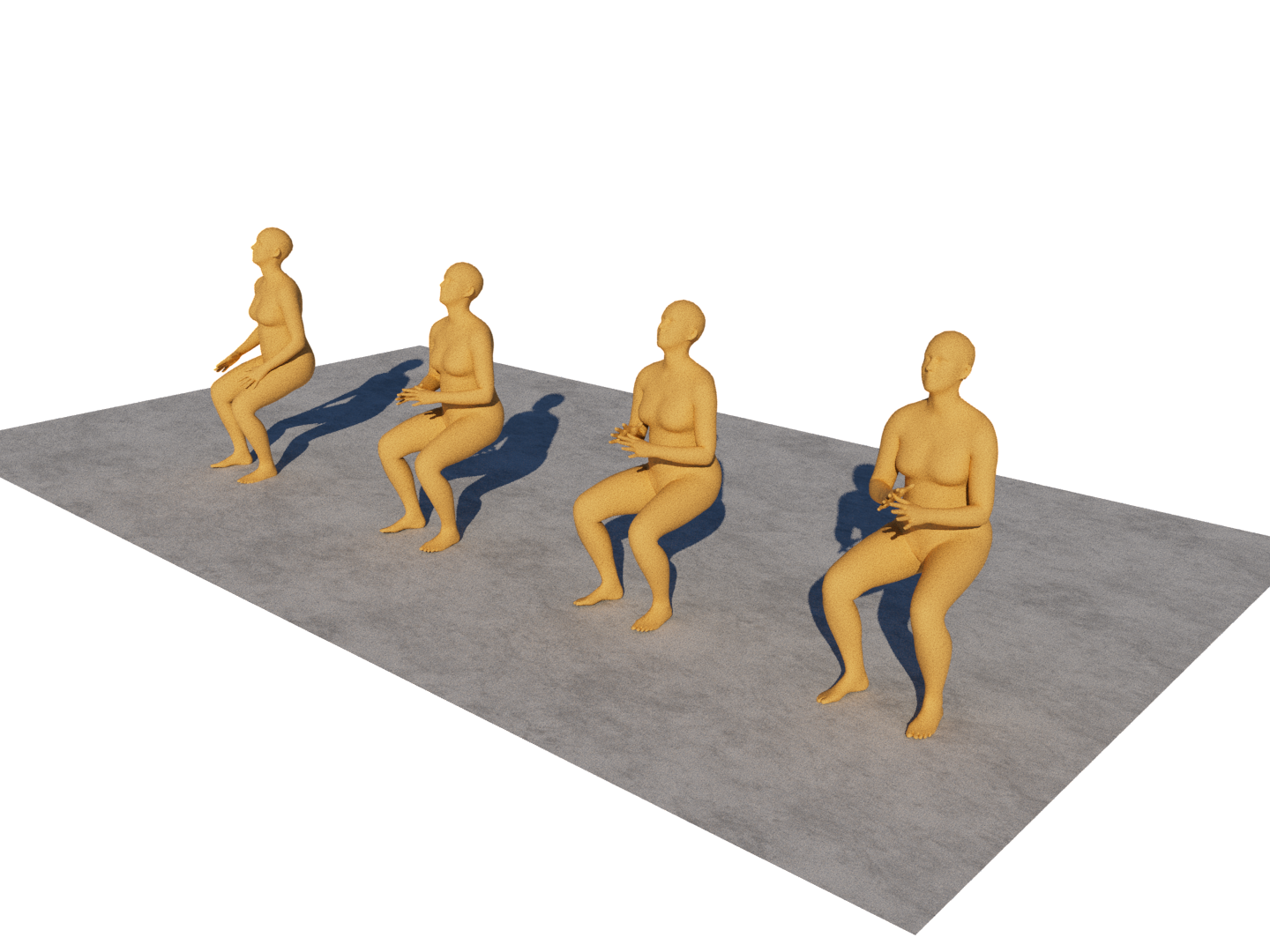} & \includegraphics[width=0.18\linewidth]{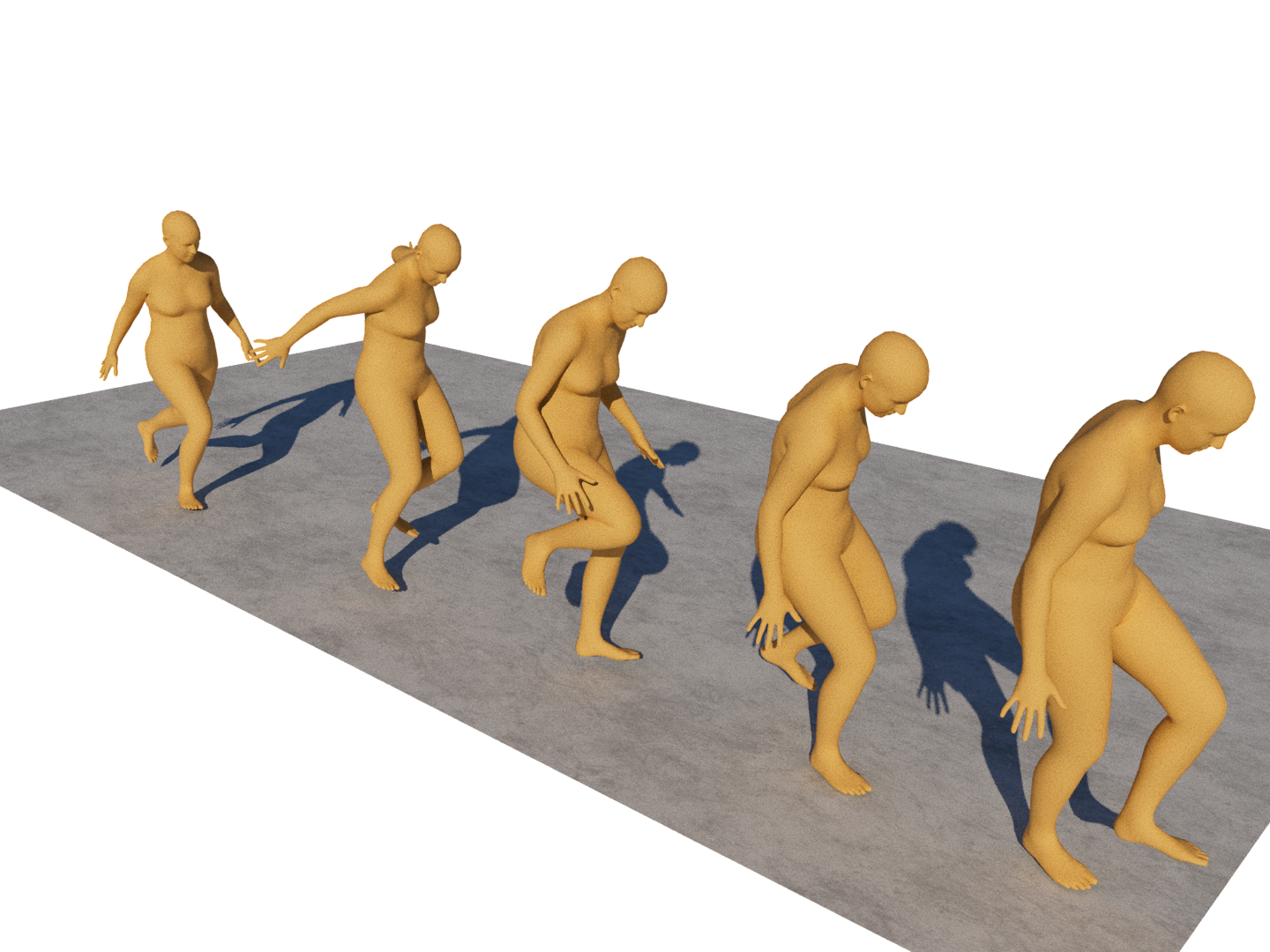}\\
    \hline
  \end{tabular}
  }
  \caption{Comparison for our KeyMotion with other multi-stage models. (A) A person is slowly tip-toeing down a path while \textbf{stretching his arms} to balance himself. (B) The person was flying around \textbf{like a fly}. (C) The figure \textbf{throws} the basketball and then \textbf{catches} it. (D) A man sits on the chair \textbf{clapping}. (E) A man is walking and \textbf{keeps jumping }to avoid something on the ground.}
  \label{fig:qualitative}
  \vspace{-3mm}
  \end{figure*}

\subsection{Qualitative Comparison}
Fig.~\ref{fig:qualitative} compares our results qualitatively with multi-stage models, T2MGPT and MLD, which are based on full motion sequences. We can see that for detailed motion descriptions, the proposed KeyMotion gives more realistic and accurate generation results. For some words like "fly like a fly", our KeyMotion presents a better understanding compared with T2MGPT and MLD. Besides, our KeyMotion generates more relevant motion sequences with sentences containing sequential changes such as "throws and catches" or accompanying motions like "sits on the chair clapping", "keeps jumping" and "while stretching his arms". 

\subsection{Comparison with Multi-Stage Reconstruction}\label{sec:multistage}

In this experiment, we remove the diffusion part and test the joint performance of VAE and MMAE. Table \ref{tab:multistage} shows that our keyframe VAE and MMAE have better reconstruction ability compared with other multi-stage generative models.  Although the VAE in the MLD paper \cite{mld} achieved the smallest FID at latent size = 7, however, their best generation results are when the latent size = 1. Such a dimension compression from $196\times263$ to $1\times256$ leads to massive information loss. Our compression from $20\times 263$ to $2\times 272$ retains more information. {For T2MGPT, their VQ-VAE {\em reconstruction} Top-3 R-Precision (Tab.~\ref{tab:multistage}) is even lower than our {\em generation's} (Tab.~\ref{tab:humanml3d}), indicating considerable information loss.}


\begin{table*}[t!]
    \renewcommand{\arraystretch}{1}
    \centering
    \caption{{Reconstruction results of the multi-stage models (MLD, T2MGPT, M2DM) on the HumanML3D dataset.} MLD's optimal latent size is 1, and the corresponding VAE is evaluated. T2MGPT's VQ-VAE applies a codebook of 512 tokens and 512 dimensions. {UNI is the uniform keyframe selection method, and MDO is the Maximum Distance Optimal keyframe selection method.}}
    \resizebox{1\textwidth}{!}{
    \begin{tabular}{lcccccc}
    \hline
    \hline
    \multirow{2}{*}{Models} & \multicolumn{3}{c}{R-precision $\uparrow$} & \multirow{2}{*}{FID$\downarrow$} & \multirow{2}{*}{MultiModal Dist$\downarrow$} &  \multirow{2}{*}{MPJPE$\downarrow$} \\
    \cline{2-4}
    & Top-1& Top-2& Top-3 & & \\ 
    \hline
    \textbf{Real} & $0.511^{\pm .003}$ & $0.703^{\pm .003}$ & $0.797^{\pm .002}$& $0.002^{\pm .000}$& $2.974^{\pm .008}$ & -\\

    \hline

    MLD-VAE-1\cite{mld} & $0.496^{\pm .003}$  & $0.689^{\pm .004}$ & $0. 784^{\pm .003}$ &  $0.247^{\pm .001}$ &  $3.076^{\pm .010}$   &${54.400}^{\pm .007}$ \\

    T2MGPT-VQ-VAE\cite{t2mgpt} & $0.501^{\pm .002}$ & $0.692^{\pm .002}$  & $0.785^{\pm .002}$ &  $0.070^{\pm .001}$& $3.072^{\pm .009}$  & $43.875^{\pm .003}$ \\

    M2DM-VQ-VAE\cite{M2DM} & $0.508^{\pm .002}$ & $0.691^{\pm .002}$  & $0.791^{\pm .002}$ &  $0.063^{\pm .001}$& $3.015^{\pm .010}$  & - \\
    
    
    MMAE + VAE (UNI) &$0.507^{\pm .004}$&$0.698^{\pm .004}$&$0.793^{\pm .003}$&$0.066^{\pm .001}$&$3.006^{\pm .009}$  & ${25.539}^{\pm .004}$ \\

    {MMAE + VAE (MDO)} &$\mathbf{0.510}^{\pm .004}$&$\mathbf{0.702}^{\pm .003}$&$\mathbf{0.796}^{\pm .003}$&$\mathbf{0.047}^{\pm .001}$&$\mathbf{3.000}^{\pm .012}$ & $\mathbf{22.141}^{\pm .005}$ \\
    \hline
    \end{tabular}
    }
    \label{tab:multistage}

    \vspace{2mm}
    \renewcommand{\arraystretch}{1}
    \centering
    \caption{Performance of different architectures for denoiser on HumanML3D test datasets.  SA stands for single attention layer and CR is condition residual. MDO is applied for all groups.}
    \resizebox{\textwidth}{!}{%
    \begin{tabular}{lcccccc}
    \hline
    \hline
    \multirow{2}{*}{Models} & \multicolumn{3}{c}{R-precision $\uparrow$} & \multirow{2}{*}{FID$\downarrow$} & \multirow{2}{*}{MultiModal Dist$\downarrow$}& \multirow{2}{*}{Diversity$\uparrow$}  \\
    \cline{2-4}
    & Top-1& Top-2& Top-3 & & &\\ 
    \hline
    \textbf{Real} & $0.511^{\pm .003}$ & $0.703^{\pm .003}$ & $0.797^{\pm .002}$& $0.002^{\pm .000}$& $2.974^{\pm .008}$ & $9.503^{\pm .065}$\\
    \hline
    SA + MLP &  $0.382^{\pm .004}$ & {$0.580^{\pm .004}$} & $0.689^{\pm .003}$ & $0.858^{\pm .005}$ & {$3.691^{\pm .022}$} & $\mathbf{9.759}^{\pm .058}$ \\
    Vanilla Transformer &  $0.407^{\pm .004}$ & {$0.603^{\pm .004}$} & $0.719^{\pm .004}$ & $0.790^{\pm .006}$ & {$3.575^{\pm .021}$}& $9.646^{\pm .054}$  \\
    CR Transformer &   $0.418^{\pm .004}$ & {$0.621^{\pm .004}$} & $0.723^{\pm .004}$ & $0.602^{\pm .008}$ & {$3.467^{\pm .024}$} & $9.583^{\pm .053}$ \\

    Skip-only Transformer &  $0.493^{\pm .004}$ & {$0.689^{\pm .003}$} & $0.784^{\pm .003}$ & $0.199^{\pm .008}$ & {$3.088^{\pm .020}$}& {$9.742^{\pm .050}$}  \\
    Parallel-only Transformer &  $0.495^{\pm .003}$ & $0.682^{\pm .003}$ & $0.782^{\pm .003}$ & $0.216^{\pm .012}$ & $3.063^{\pm .019}$ & $9.714^{\pm .064}$\\
    PST (ours) &  {$\mathbf{0.499}^{\pm .003}$} & {$\mathbf{0.700}^{\pm .003}$} & {$\mathbf{0.794}^{\pm .003}$} & {$\mathbf{0.192}^{\pm .008}$} & {$\mathbf{3.036}^{\pm .010}$}  & $9.803^{\pm .100}$ \\
    \hline
    \end{tabular}
    }

    \label{tab:denoiser}
\end{table*}

\section{Ablation Studies}


\noindent \textbf{PST Denoiser.} 
We applied multiple architectures and compared their performance with standard metrics. Table \ref{tab:denoiser} shows that single attention with MLP (row 2) and the Vanilla stacked Transformer encoder (row 3) are not capable of learning the reverse diffusion process effectively. 
With conditional residual added to each layer (row 4), the performance improves but still does not match the Skip-only Transformer (row 5). 
For a fair comparison, we doubled the dimension of the Skip-only Transformer and the number of attention heads to maintain a similar parameter size to PST. 
Our parallel-only Transformer (row 5) achieves compatible performance with the Skip-only Transformer. The proposed Parallel Skip Transformer (last row) outperforms all other configurations. {A possible explanation is that conventional stacked Transformers suffer from gradient vanishing without layer-level skip connections. On the other hand, the parallel Transformer with a two-stream structure gives an alternative "highway" for the gradient to descend.}

\begin{table}[t!]
\centering
\caption{FID, top-1 R-precision, Multimodal Distance, MPJPE, and PAMPJPE in millimeter for unconditional MMAE and conditional MMAE with different levels of noise added to the 10\% keyframes on the HumanML3D dataset. GT is ground truth full sequence. The last two rows show the result of motion infilling with generated keyframes and, therefore have no MPJPE and PAMPJPE performance.}
\resizebox{0.48\textwidth}{!}
{
  \begin{tabular}{clccccr}
    \hline
    \hline
    Noise&Method & FID & RP Top-1 & MMD &  PAMPJPE &  MPJPE   \\
    \hline
    \multirow{3}{*}{0.3 $\mathcal{N}(0,1)$}&GT&0.820&0.475&3.106 & 54.030  & 69.593\\
    &uncond &$0.061$&$0.505$&2.996 & 17.137 & 34.446\\
    &cond&0.060 &0.508&2.997 &  15.901 & 28.015\\ 
    \hline
    \multirow{3}{*}{0.5 $\mathcal{N}(0,1)$}&GT &3.209&0.427&3.517 & 88.913  & 115.891\\
    &uncond&0.763 &0.471&3.131 & 28.174 & 52.676\\
    &cond &0.541&0.475&3.105 & 24.172 & 43.159\\
    \hline
    \multirow{3}{*}{0.7 $\mathcal{N}(0,1)$}&GT &7.336&0.378&4.092 & 122.23 & 162.320\\
    &uncond&3.269&0.423&3.567 & 40.109 & 66.217\\
    &cond &2.025& 0.438&3.395 & 32.561 & 58.739 \\
    \hline
    Generated &uncond & 0.245 &0.482&3.125 & - & - \\
     Keyframe &cond&0.192&0.499&3.036 & - & -\\
    \hline
  \end{tabular}
  \label{table:uncond}
}
\end{table}

\noindent \textbf{Text Conditioning in MMAE.}
Table \ref{table:uncond} illustrates the significance of text conditioning in MMAE. We also use mean per joint position error (MPJPE) in millimeters in this experiment, which indicates the distance between reconstructed motion and ground truth motion. Comparing the reconstruction ability with different Gaussian noise levels, our MMAE with text conditioning shows higher robustness to noise giving better infilling results. 

\vspace{1mm}
\noindent \textbf{Additional Results.} 
In the supplementary material, we evaluate the MMAE performance with different numbers of layers and find that 8 layers give the best reconstruction performance. 
We also give additional results by studying the effects on latent sizes, the classifier-free guidance factor, and different keyframe selection strategies in the supplementary material.


\vspace{-3mm}
\section{Conclusion}
    
We proposed KeyMotion, a technique for generating 3D human motion driven by text. Our method first generates keyframes in latent space using a novel Parallel Skip Transformer that performs cross-modal attention between the keyframe latents and text conditions. We also proposed a Motion Masked AutoEncoder based on a text-conditioned residual Transformer that completes the 3D motion given keyframes while preserving the motion fidelity and ensuring adherence to the physical human body constraints. Extensive experiments on two benchmark datasets and ablation studies show the efficacy of our proposed method.

\bibliographystyle{splncs04}
\bibliography{keymotion}

\section*{Acknowledgments}
This should be a simple paragraph before the References to thank those individuals and institutions who have supported your work on this article.


 




\newpage

\appendix

\subsection{Ablation Study on Keyframe Selection Method}

We also included other two keyframe selection methods, Minimum Interpolation Error Optimal Keyframe Selection \cite{roberts2018optimal} and Interval Maximum Distance Optimal Keyframe Selection. The former is defined as the sequence with minimum accumulated interpolation by maintaining a minimum interpolation error:

\begin{equation}
    E_{i,j}^m = \min_k E_{i,k}^{m-1} + e_{k,j},
\end{equation}

where $E_{i,k}^m$ is the minimum accumulated interpolation error from $i$-th frame and $j$-th frame, through $m$ frames, and $e_{k,j}$ is the Mean Square Error between the interpolation estimation and the ground truth sequence the from the $k$-th frame to the $j$-th frame. 

The latter method defines keyframes as the sequences that find the sub-sequence $\{Kf\}_{i=1}^K$ with maximum distance using the greedy algorithm $Kf_{i+1} = F[\arg \max_k || F_k - KF_{i} ||_2 ]$ in uniform intervals $I_{i+1}$, where $F_i$ is the $i$-th frame in the motion. 

\vspace{-1mm}
\begin{table*}[h]
    \centering
    \caption{KeyMotion performance on the \textbf{HumanML3D} with different keyframe selection strategies. IMDO stands for Interval Maximum Distance Optimization, MIEO stands for Minimum Interpolation Error Optimization, and MDO stands for Maximum Distance Optimization. MPJPE between ground truth keyframes is also calculated to illustrate the proximity.}
    \resizebox{1\textwidth}{!}{%
    \begin{tabular}{lcccccc}
    \hline
    \multirow{2}{*}{Keyframe Selction} & \multicolumn{3}{c}{R-precision $\uparrow$} & \multirow{2}{*}{FID$\downarrow$} & \multirow{2}{*}{MultiModal Dist$\downarrow$} & \multirow{2}{*}{MPJPE}\\
    \cline{2-4}
     & Top-1& Top-2& Top-3 & & &\\ 
    \hline
    \textbf{Real} & $0.511^{\pm .003}$ & $0.703^{\pm .003}$ & $0.797^{\pm .002}$& $0.002^{\pm .000}$& $2.974^{\pm .008}$ & $13.201^{\pm .003}$\\

    Uniform &  $0.494^{\pm.004}$ & $0.688^{\pm.004}$ & $0.786^{\pm.003}$ & $0.178^{\pm.010}$ & $3.083^{\pm.008}$&$88.765^{\pm.005}$\\

    IMDO & $0.493^{\pm.004}$ & $0.686^{\pm.003}$ & $0.788^{\pm.003}$ & $\mathbf{0.162^{\pm.010}}$ & $3.067^{\pm.010}$&$90.910^{\pm.005}$\\

    MIEO & $0.498^{\pm.003}$ & $0.692^{\pm.004}$ & $0.791^{\pm.003}$ & $0.169^{\pm.010}$ & $3.045^{\pm.011}$&$92.384^{\pm.005}$\\

    MDO& $\mathbf{0.499^{\pm.003}}$ & $\mathbf{0.700^{\pm.003}}$ & $\mathbf{0.795^{\pm.003}}$ & $0.192^{\pm.008}$ & $\mathbf{3.036^{\pm.010}}$ & $94.732^{\pm.005}$\\
    \hline
    \end{tabular}
    }

    \label{tab:kf}
\end{table*}

Apart from the performance of generation results, the MPJPEs for all keyframe selection methods are also demonstrated.
We found that with the decreasing proximity between adjacent frames, the performance of the generation result increases in terms of motion-text alignment.

\subsection{Ablation Study on Number of Layers of MMAE }

We conduct experiments for different numbers of Transformer Layers of MMAE and evaluate the completing ability on the test dataset in terms of FID, top-1 R-precision, and MultiModal Distance.

\begin{table}[htp]
  \caption{Motion completion performance of MMAE with different layers. Each conditional MMAE is trained by 1000 epochs.}
\centering
  \begin{tabular}{lcccc}
    \hline
    Method & FID & RP Top-1 & MultiModal Dist \\
    \hline

    cond 4 layer &0.024&0.509&3.004\\
    cond 5 layer &0.022&0.509&3.000\\
    cond 6 layer & 0.029 &{0.510}&{3.001}\\
    cond 7 layer & 0.014 &{0.511}&{3.002}\\
    cond 8 layer &\textbf{0.010}&\textbf{0.510}&\textbf{3.000}\\
    cond 10 layer &{0.019}&{0.510}&{3.000}\\
    cond 12 layer &{0.026}&{0.503}&{3.010}\\
    \hline
  \end{tabular}

  \label{tab:layernummae}
\end{table}


\subsection{Latent Size}
We also explored the performance with different latent sizes $s_l$ shown in Tab~\ref{tab:latentsize}. We found that the VAE and MMAE achieved the best reconstruction result at latent size=3, but when the latent size is 2, the whole approach achieves the best performance for generation.

\begin{table}[thp]
    \renewcommand{\arraystretch}{1.2}
    \centering
    \caption{KeyMotion performance on the \textbf{HumanML3D} with different latent dimension $d_l$. The latent dimension for all keyframe rate is 2 and the keyframe for all latent dim groups are 10\%}
    \begin{tabular}{lccccc}
    \hline
    \multirow{2}{*}{Latent Size} & \multicolumn{3}{c}{R-precision $\uparrow$} & \multirow{2}{*}{FID$\downarrow$} & \multirow{2}{*}{MultiModal Dist$\downarrow$} \\
    \cline{2-4}
     & Top-1& Top-2& Top-3 & &  \\ 
    \hline
    $s_l=1$ &  $0.477^{\pm .003}$ & {$0.676^{\pm .003}$} & $0.772^{\pm .003}$ & $0.361^{\pm .008}$ & {$3.163^{\pm .021}$} \\
    $s_l=2$ &   $\mathbf{0.499^{\pm .003}}$ & $\mathbf{0.700^{\pm .003}}$ & $\mathbf{0.794^{\pm .003}}$ & $\mathbf{0.192^{\pm .008}}$ & $\mathbf{3.036^{\pm .010}}$ \\
    $s_l=3$ &  {$0.465^{\pm .003}$} & $0.675^{\pm .003}$ & $0.774^{\pm .003}$ & $0.293^{\pm .009}$ & $3.171^{\pm .019}$ \\
    \hline
    \end{tabular}

    \label{tab:latentsize}
\end{table}

\begin{table}[h!]
\renewcommand{\arraystretch}{1.22}
\centering
  \caption{Reconstruction performance of MMAE  and VAE with different latent sizes. The best reconstruction metrics are marked in bold font.}
  \begin{tabular}{lccc}
    \hline
    Latent Size & FID & RP Top-1 & MM Distance \\
    \hline

    $s_l=$ 1 &0.102&0.500&3.018\\
    $s_l=$ 2 &0.047&0.510&3.000\\
    $s_l=$ 3 &\textbf{0.042}&\textbf{0.510}&\textbf{2.995}\\
    \hline
  \end{tabular}
 \label{tab:latentdimvae}

\end{table}

\subsection{Parallel Skip Transformer Convergence}

\begin{figure}[h!]
  \centering
    \includegraphics[width=0.48\textwidth]{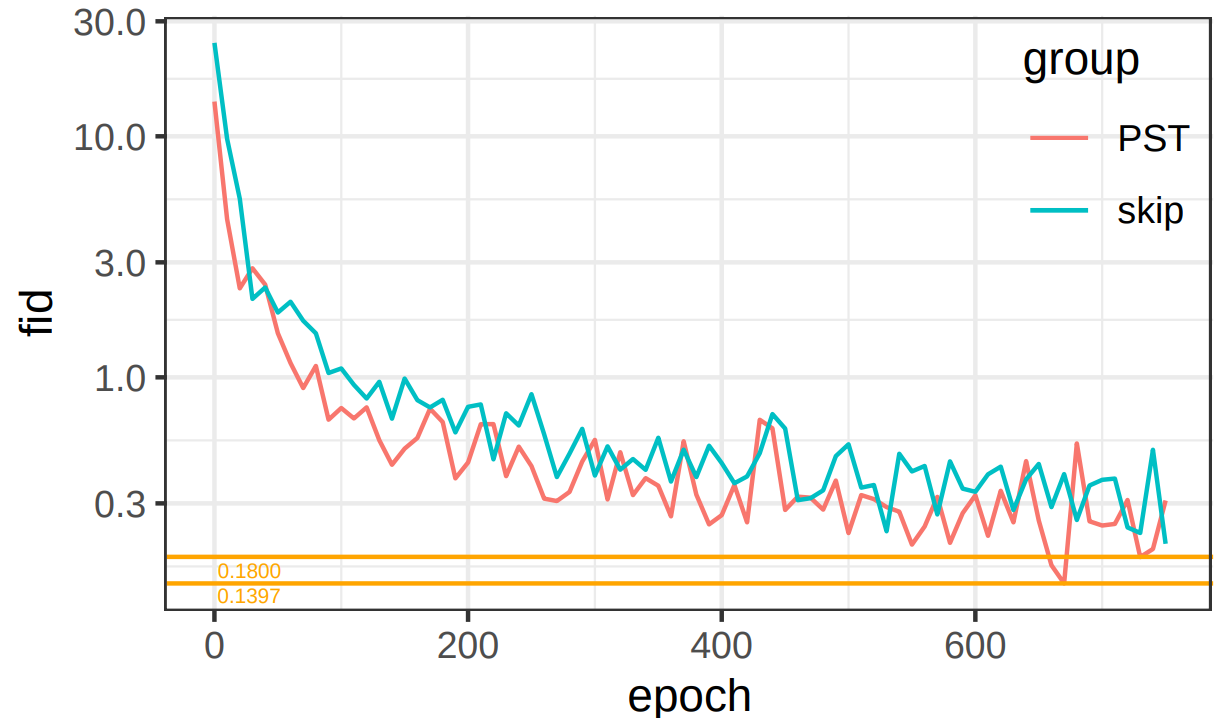}
    \centering
    \caption{FID over epochs}
    \label{fig:fidpstskip}
    \includegraphics[width=0.48\textwidth]{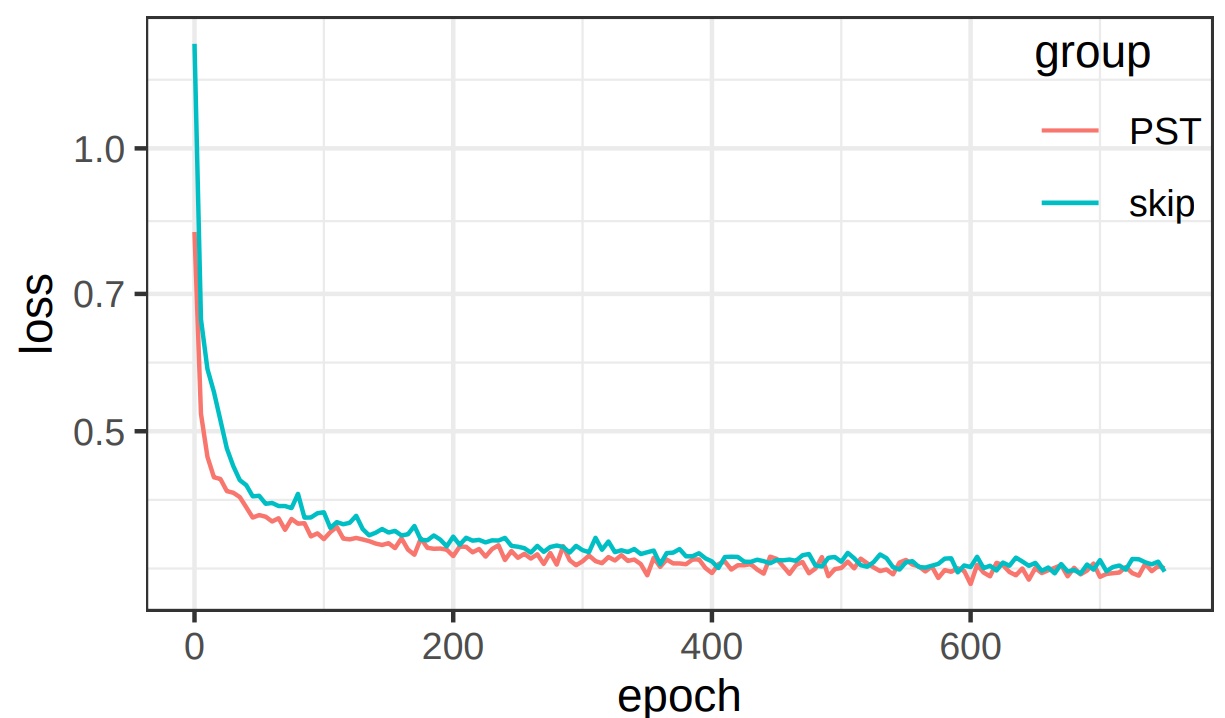}
    \centering
    \caption{Denoise loss over epochs}
    \label{fig:losspstskip}
  \caption{Performance of PST and Skip-only Transformer in terms of FID and denoise loss on the HumanML3D test dataset.}
  \label{fig:pstvsskip}
\end{figure}

The proposed PST has a higher efficiency in learning the reverse diffusion process for text-to-motion tasks. Compared to the skip-only Transformer, the PST converges more quickly at a lower denoising loss. We trained the PST and skip-only Transformer for 750 epochs with a constant learning rate of $5\times 10^{-5}$ and evaluated them on the test dataset to monitor the performance for every 10 epochs. The plot in Fig~\ref{fig:pstvsskip}
shows that PST outperforms Skip-Transformer in both FID and loss. 

\subsection{Classifier Free Guidance Factor}

\begin{figure}[h!]
\centering
    \includegraphics[width=\linewidth]{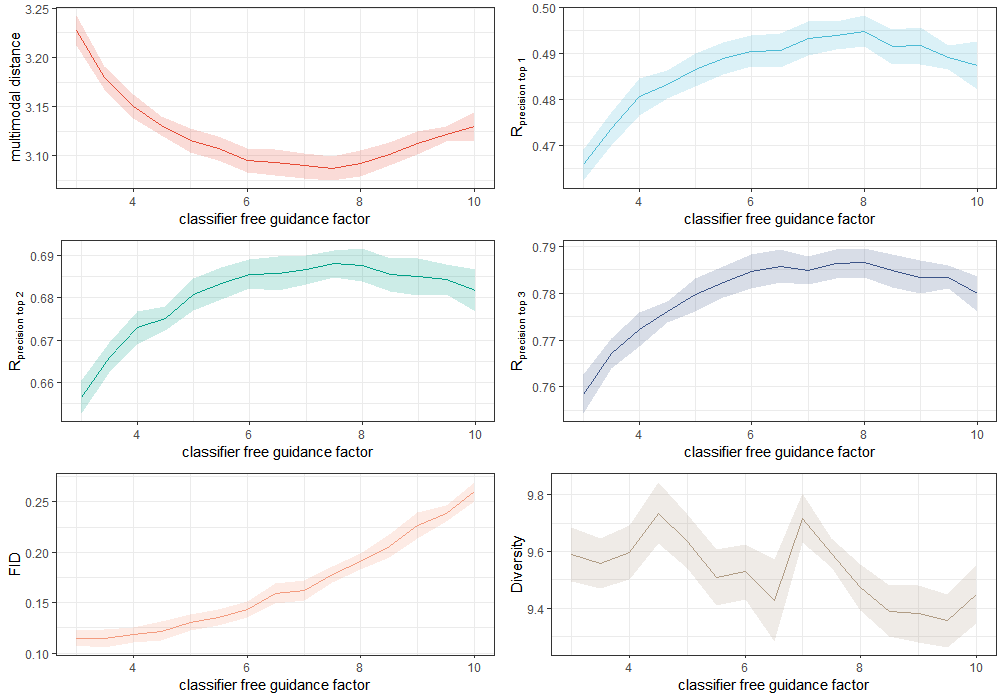}
    \caption{Metrics change over classifier-free guidance factor.}
    \label{fig:cfgsweep}
\end{figure}

We explore the metrics changes over the classifier-free guidance factor $s$, finding that the FID increases over $s$ but a trade-off is made for R-precisions and MultiModal Distance shown in Fig.~\ref{fig:cfgsweep}. 

\subsection{Average Inference Time per Sentence}

We calculated our average inference time per sentence on an RTX 3090 GPU, and the result is 0.34 s, which is capable of real-time generation. 

\subsection{Bone Length Constraints in MMAE}

Our experiments suggest that our MMAE with pure MSE loss for the redundant motion representation shows similar quantitative results to the one with bone length constraints (BLC) but causes serious noise and joggle. This issue is efficiently solved with the constraints on bone length for both the keyframe VAE and MMAE. The corresponding animation with BLC and without BLC is presented in the same MP4 file as the animation results.

\subsection{Combining MMAE with Other Models}

We evaluate the efficiency of our MMAE by picking 
the keyframes (to match our method) from the full sequence generated by other models. These keyframes are then passed to our MMAE.

\begin{table*}[h]
    \centering
    \caption{MMAE combined with other models on \textbf{HumanML3D} dataset.}
    \resizebox{1\textwidth}{!}{%
    \begin{tabular}{lcccccc}
    \hline
    \multirow{2}{*}{Keyframe Selction} & \multicolumn{3}{c}{R-precision $\uparrow$} & \multirow{2}{*}{FID$\downarrow$} & \multirow{2}{*}{MultiModal Dist$\downarrow$} & \multirow{2}{*}{Multimodality}\\
    \cline{2-4}
     & Top-1& Top-2& Top-3 & & &\\ 
    \hline
    \textbf{Real} & $0.511^{\pm .003}$ & $0.703^{\pm .003}$ & $0.797^{\pm .002}$& $0.002^{\pm .000}$& $2.974^{\pm .008}$ & -\\

    MDM &  $0.418^{\pm.005}$ & $0.604^{\pm.005}$ & $0.708^{\pm.004}$ & $0.490^{\pm.025}$ & $3.645^{\pm.028}$&$2.873^{\pm.111}$\\

    MDM + MMAE &  $0.420^{\pm.005}$ & $0.600^{\pm.004}$ & $0.710^{\pm.004}$ & $0.437^{\pm.017}$ & $3.608^{\pm.018}$&$2.549^{\pm.127}$\\

    MLD &  $0.481^{\pm.003}$ & $0.673^{\pm.003}$ & $0.772^{\pm.002}$ & $0.473^{\pm.013}$ & $3.196^{\pm.010}$&$2.413^{\pm.079}$\\

    MLD + MMAE &  $0.477^{\pm.003}$ & $0.668^{\pm.003}$ & $0.769^{\pm.003}$ & $0.388^{\pm.010}$ & $3.214^{\pm.008}$&$2.211^{\pm.084}$\\

    \hline

    \end{tabular}
    }

    \label{tab:mmae}
\end{table*}

From Tab.\ref{tab:mmae} we can see that our MMAE significantly improves the FID of other models, but is unable to 
improve the R-precision values and the MultiModal Distance. This shows the significance of explicitly generating only the keyframes as done by our KeyMotion method. For MDM, we discovered that shuffling the generated dataset enhances performance. Consequently, we assessed the performance of MDM, whose original code does not shuffle, using the shuffled-generated results.

\subsection{Model Size and Efficiency}
\begin{table}[htp]
    \centering
    \caption{The Parameter size and FLOPs for single text input of our model and other diffusion-based models}
    \begin{tabular}{cccc}
    \hline
    Model  & \quad Parameters \quad & \quad FLOPs \quad & \quad FID \quad \\
    \hline
        MDM  & 34.6M & 7.3 T & 0.544\\
        MotionDiffuse & 80.0M & 7.1 T & 0.630 \\
        MLD  & \textbf{26.3M} & 15.8 B & 0.473\\
        KeyMotion(Ours)  & 29.0M & \textbf{15.5 B} & \textbf{0.192}\\
    \hline
    \end{tabular}

    \label{tab:sizeeff}
\end{table}
We compared our model together with other diffusion-based models on model size and efficiency together with FID. For fairness, all the models are compared excluding the parameter number of the CLIP text encoder.
Tab. \ref{tab:sizeeff} suggests that our model has much fewer parameters than MDM and MotionDiffuse, it also involves the fewest floating-point operations throughout the entire generation process \textbf{including} motion completing. Therefore our latent diffusion model is not only compatible with parameter numbers but also achieves high inference efficiency for real-time demand. It's noteworthy that despite our larger total parameter number being slightly higher compared to MLD, our overall FLOPs are lower. This discrepancy arises from the denoising process consuming the majority of inference time, wherein our model boasts fewer parameters in both VAE and the denoiser.

\subsection{Dataset Descriptions}
   \noindent\textbf{HumanML3D} dataset uses a redundant motion representation of 263 dimensions for each frame with 22 joints (key points). The 263 dimensions are a tuple of $(\dot{r}^a,\dot{r}^x,\dot{r}^z,r^y,\mathbf{j}^p,\mathbf{j}^v,\mathbf{j}^r, \mathbf{c}^f) $, where $\dot{r}^a \in \mathbb{R}$ is root angular velocity along Y-axis; $\dot{r}^x,\dot{r}^z \in \mathbb{R}$ are
    root linear velocities on XZ-plane; $r^y \in \mathbb{R}$ is root height; $\mathbf{j}^p,\mathbf{j}^v,\mathbf{j}^r \in \mathbb{R}^{3j}, \mathbb{R}^{3j}, \mathbb{R}^{6j}$ are the local joints positions, velocities and rotations in root space, with j denoting
    the number of joints; $c^j \in \mathbb{R}^4$ is the foot contact label. 
    \noindent\textbf{KIT-ML}: dataset uses a similar motion representation to HumanML3D, but it has 21 joints and thus 251 dimensions.

\subsection{Details of Evaluation}

    \noindent \textbf{FID}: Frechet Inception Distance is used to evaluate the dissimilarity between two distributions as  
    \begin{align}
         &d_F( \mathcal{N}(\mathbf{\mu},\mathbf{\Sigma}) , \mathcal{N}(\mathbf{\mu}',\mathbf{\Sigma}'))^2 = \notag \\ 
       &||\mathbf{\mu}-\mathbf{\mu}'||^2 + Tr(\mathbf{\Sigma} + \mathbf{\Sigma}' - 2\sqrt{\mathbf{\Sigma}\mathbf{\Sigma}'}), 
    \end{align}

    \noindent where the $\mathbf{\mu} \text{ and } \mathbf{\mu}'$ are the feature's mean values of generated samples and ground truth, and $\mathbf{\Sigma} \text{ and } \mathbf{\Sigma}'$ are the feature's covariance matrices respectively. This metric measures the distance between two normal distributions, that is generated samples and the ground truth. The less the FID is, the better the model performs.
       Since generated motion results usually contain hundreds or even thousands of frames, FID for motion samples should be calculated with extracted features instead of raw generated motion data and ground truth. In practice, pre-trained action recognition models by Guo et al. \cite{a2m} and by Ji et al. \cite{uestc} are utilized to extract features from input samples.
   
    \noindent \textbf{Diversity}: Diversity measures the variance of the generated motions across all action categories. From a set of all generated motions from various action types, two subsets of the same size $S_d$ are randomly sampled, and their respective feature sets are extracted as $\{\mathbf{f_1},\cdots,\mathbf{f_{S_d}}\}$ and $\{\mathbf{f'_1},\cdots,\mathbf{f'_{S_d}}\}$. The diversity between them is given by:
    \begin{equation}
        \text{Div} = \frac{1}{S_d}\sum_{i=1}^{S_d}||\mathbf{f_i} - \mathbf{f'_i}||^2
    \end{equation}
    Usually, good generated samples are supposed to have a similar diversity value as ground truth. 
    \noindent \textbf{Multimodality}: Different from diversity, multimodality measures how much the generated motions diversify within each action type. Given a set of motions with $C$ action types. For $c$-th action, we randomly sample two subsets with same size $S_l$ , and then extract two subset of feature vectors $\{\mathbf{f}_{c,1}, ... , \mathbf{f}_{c,S_l} \}$ and $\{\mathbf{f}'_{c,1}, ... , \mathbf{f}'_{c,S_l} \}$. The multimodality of this motion set is formalized as
    \begin{equation}
        \text{Multimodality} = \frac{1}{C\cdot S_l}\sum_{c=1}^{C}\sum_{i=1}^{S_l}|| \mathbf{f_{c,i}} - \mathbf{f'_{c,i}} ||^2
    \end{equation}
    \noindent \textbf{Multimodal Distance}: This metric describes how close the relationship is between the motion features and the text features. It is computed as the average Euclidean distance between the generated motion features and corresponding text features. For given text $\mathbf{T}$ with feature $\mathbf{f_T}$ and corresponding generated motion $\mathbf{M}$ with feature $\mathbf{f_M}$. The multimodal distance is given by:
    \begin{equation}
        \text{MMD} = \sqrt{\frac{1}{n}\sum_{i=1}^{n}||\mathbf{f_{T,i}} - \mathbf{f_{M,i}}||^2}
    \end{equation}
    where $n$ is the sample number of text $\mathbf{T}$. In practice, the text features are extracted by a pre-trained text encoder by Guo et al. \cite{text2motion}. Multimodal distance is also applied for motion generation tasks driven by other modalities than text, such as music and audio. 
    
    \noindent \textbf{R-Precision}: Also known as motion-retrieval precision, it calculates the text and motion's top $K$ matching accuracy among $R$ documents based on the Euclidean distance. Actually, most works follow the method of Guo et al. \cite{text2motion}. For each generated motion, its ground-truth text description and 31 randomly selected mismatched descriptions from the test set form a description pool. This is followed by calculating and ranking the Euclidean distances between the motion feature and the text feature of each description in the pool. Then accuracy will be calculated for the top 1, 2, and 3 places.

\subsection{More Results}
\begin{figure*}[htp]
\centering
    \includegraphics[width=0.8\linewidth]{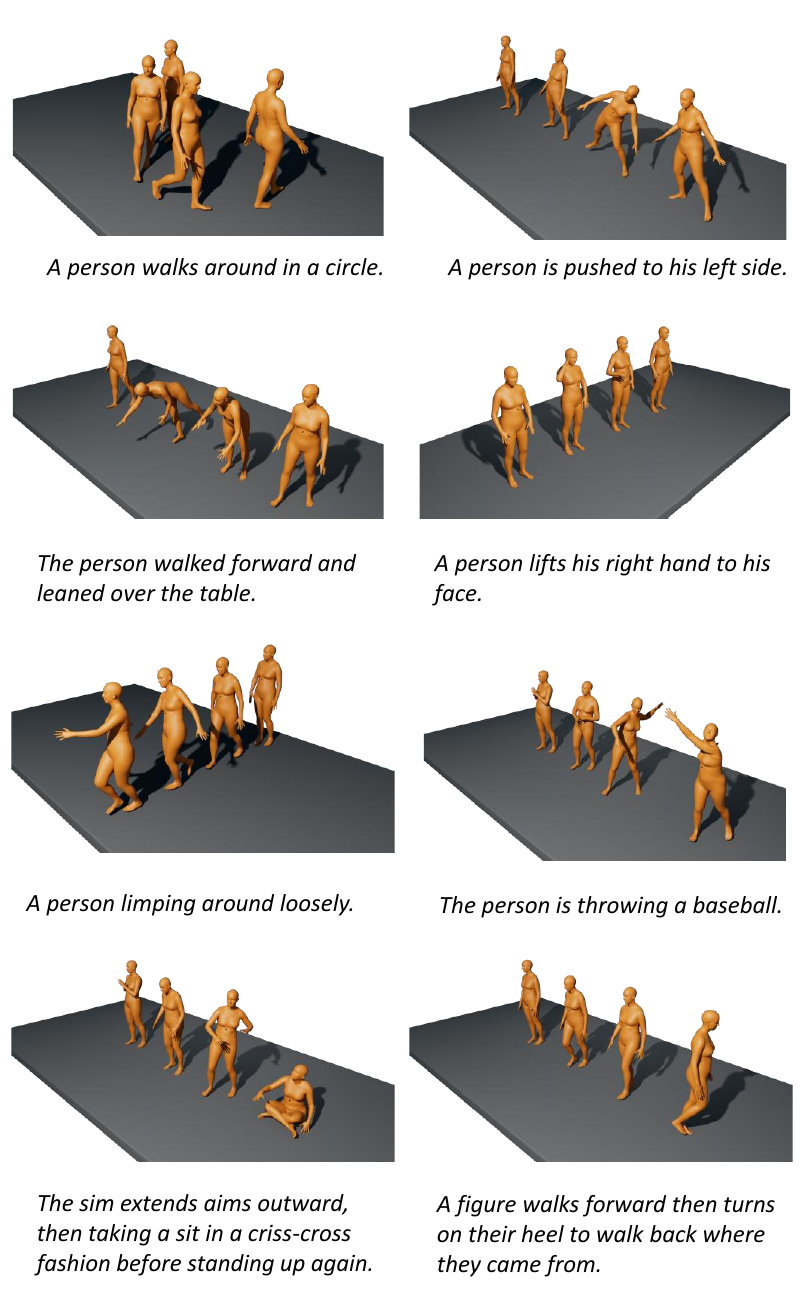}
    \caption{Motion Results 1}
    \label{fig:gallery1}
\end{figure*}

\begin{figure*}[htp]
\centering
    \includegraphics[width=0.8\linewidth]{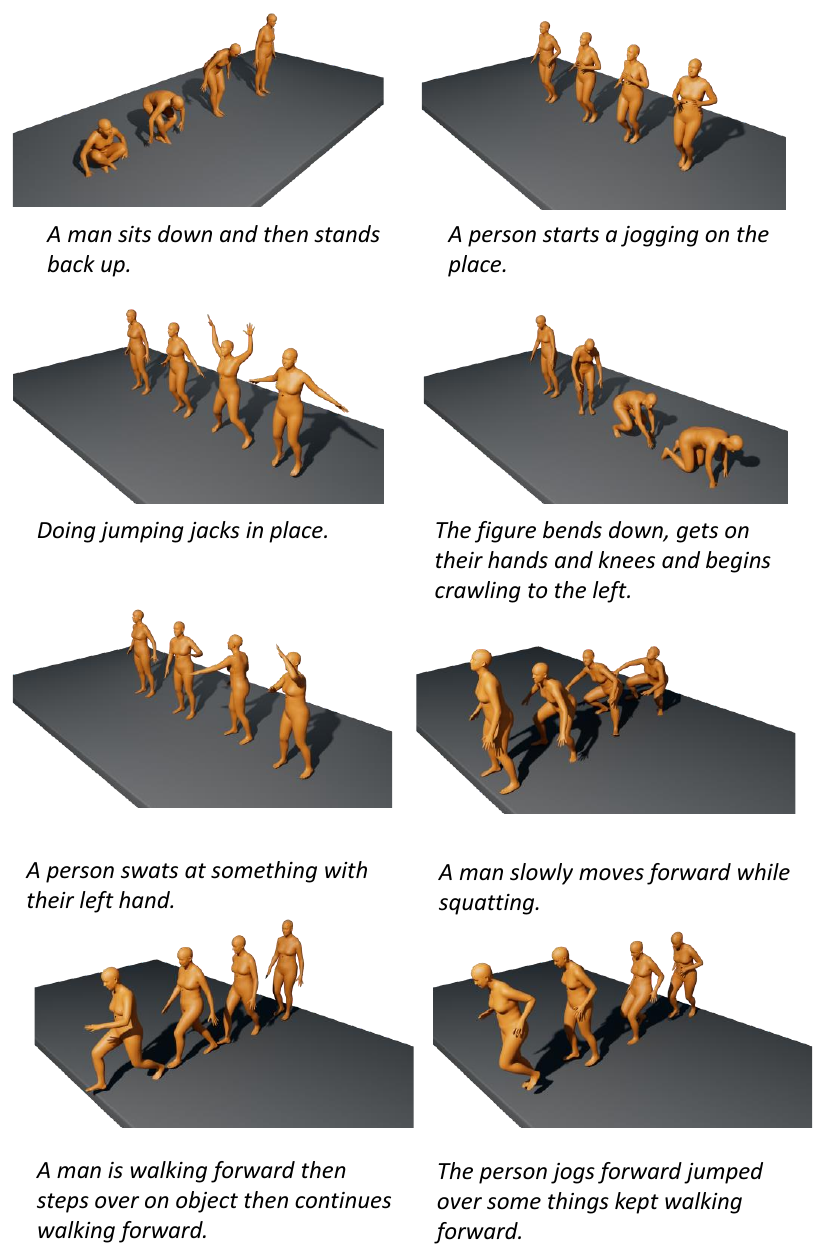}
    \caption{Motion Results 2}
    \label{fig:gallery2}
\end{figure*}

We present various generated motion sequences and the keyframes are shown in Fig~\ref{fig:gallery1} and \ref{fig:gallery2}, whose corresponding animations are attached in an MP4 file named "supplementary.mp4". Please also watch the animations provided in the MP4 file for better visualizations. 

\vfill

\end{document}